\pdfoutput=1
\documentclass[11pt]{article}
\usepackage[preprint]{acl}
\usepackage{times}
\usepackage{latexsym}
\usepackage[T1]{fontenc}
\usepackage[utf8]{inputenc}
\usepackage{microtype}
\usepackage{inconsolata}
\usepackage{xcolor}
\usepackage{graphicx}
\usepackage{booktabs}
\usepackage{subcaption}
\usepackage{colortbl}
\definecolor{light-light-gray}{gray}{0.92} 
\usepackage{listings}
\usepackage{tcolorbox}
\usepackage{multirow}
\usepackage{fancyvrb}
\usepackage{xcolor}
\usepackage{tcolorbox}
\tcbuselibrary{skins}
\usepackage{fancyvrb}

\definecolor{promptbackground}{RGB}{248, 250, 252}
\definecolor{promptborder}{RGB}{203, 213, 225}
\definecolor{prompttitle}{RGB}{15, 23, 42}
\definecolor{codebackground}{RGB}{241, 245, 249}

\newtcolorbox{promptbox}{
  colback=promptbackground,
  colframe=promptborder,
  fonttitle=\bfseries\color{white},
  title=GPT-4o Negation Prompt,
}

\definecolor{errorbackground}{RGB}{254, 242, 242}
\definecolor{errorborder}{RGB}{248, 113, 113}
\definecolor{errortitle}{RGB}{153, 27, 27}
\definecolor{originalcolor}{RGB}{34, 197, 94}
\definecolor{negatedcolor}{RGB}{239, 68, 68}

\newtcolorbox{errorbox}{
  colback=errorbackground,
  colframe=errorborder,
  fonttitle=\bfseries\color{white},
  title=GPT-4o Negation Errors,
}

\newtcolorbox{examplebox}[1]{
  colback=white,
  colframe=gray!30,
  top=2mm,
  bottom=2mm,
  left=3mm,
  right=3mm,
  before skip=3mm,
  after skip=3mm,
  title=#1,
  fonttitle=\small\bfseries
}

\title{\emph{NegVQA}: Can Vision Language Models Understand Negation?}

\author{
 \textbf{Yuhui Zhang\textsuperscript{1}}\quad\quad
 \textbf{Yuchang Su\textsuperscript{2}}\quad\quad
 \textbf{Yiming Liu\textsuperscript{2}}\quad\quad
 \textbf{Serena Yeung-Levy\textsuperscript{1}}
\\
 \textsuperscript{1}Stanford University\quad\quad
 \textsuperscript{2}Tsinghua University
\\
 \small{
   \textbf{Correspondence:} \href{mailto:yuhuiz@stanford.edu}{yuhuiz@stanford.edu}
 }
}

\begin{document}
\maketitle

\begin{abstract}
Negation is a fundamental linguistic phenomenon that can entirely reverse the meaning of a sentence. As vision language models (VLMs) continue to advance and are deployed in high-stakes applications, assessing their ability to comprehend negation becomes essential. To address this, we introduce \emph{NegVQA}, a visual question answering (VQA) benchmark consisting of 7,379 two-choice questions covering diverse negation scenarios and image-question distributions. We construct \emph{NegVQA} by leveraging large language models to generate negated versions of questions from existing VQA datasets. Evaluating 20 state-of-the-art VLMs across seven model families, we find that these models struggle significantly with negation, exhibiting a substantial performance drop compared to their responses to the original questions. Furthermore, we uncover a U-shaped scaling trend, where increasing model size initially degrades performance on \emph{NegVQA} before leading to improvements. Our benchmark reveals critical gaps in VLMs' negation understanding and offers insights into future VLM development. Project page available at \url{https://yuhui-zh15.github.io/NegVQA/}.
\end{abstract}

\section{Introduction}

\begin{figure*}
    \centering
    \includegraphics[width=\linewidth]{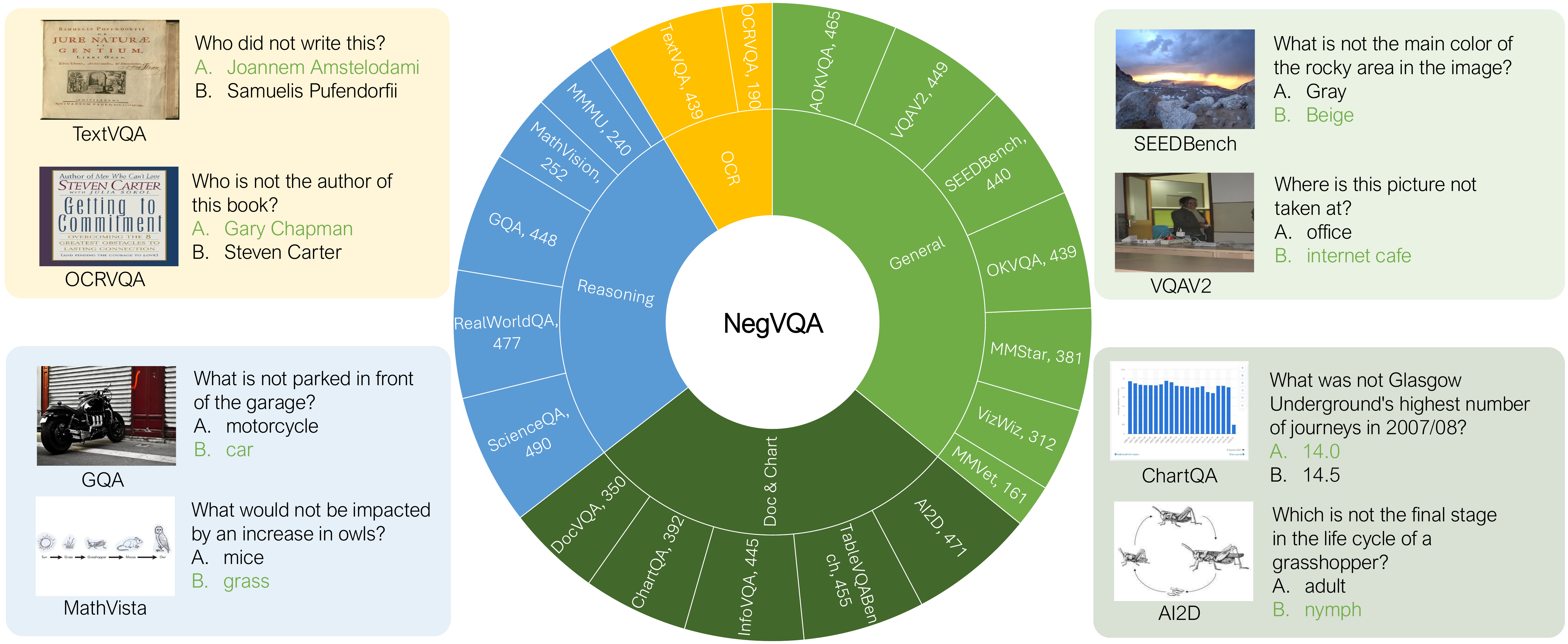}
    \caption{\textbf{\emph{NegVQA} dataset overview.} \textit{(Middle)} \emph{NegVQA} comprises a diverse set of negated questions, totaling 7,379 instances sourced from various VQA datasets and domains (general, document/chart, reasoning, OCR). \textit{(Left/Right)} Example questions from different datasets and domains, with correct answers highlighted in green. }
    \label{fig:dataset}
\end{figure*}

Vision language models (VLMs) such as GPT-4o and Claude have demonstrated remarkable capabilities in understanding and reasoning about visual content through natural language interactions~\cite{gpt4,claude}. These models can answer image-based questions, generate descriptions, and engage in multi-turn dialogues about visual scenes~\cite{llava1.5,deitke2024molmo,qwen2vl}. More recently, they have been integrated into embodied AI systems and robotics, allowing direct interaction with environments and humans in high-stakes scenarios~\cite{driess2023palm,brohan2023rt,kim2024openvla}.

Despite their impressive progress, VLMs’ ability to understand negation~\cite{ackrill1975categories}—a fundamental linguistic phenomenon that can completely alter the meaning of a sentence—remains poorly understood. A failure to correctly interpret negation can lead to critical errors, particularly in interactive AI systems. For instance, if a user instructs a VLM not to take a certain action or asks about something that is absent, misunderstanding negation could result in actions contrary to user intent and pose serious safety risks.

To address this, we introduce \emph{NegVQA}, a visual question answering (VQA) benchmark designed to assess VLMs’ comprehension of negation. While existing VQA datasets primarily focus on affirmative questions, \emph{NegVQA} systematically examines negation understanding across diverse scenarios. The dataset consists of 7,379 two-choice questions, covering a range of negation types, including cases where objects are absent, attributes such as colors or sizes are negated, actions are described in terms of what is not happening, and more complex forms of negation that require deeper reasoning. To construct \emph{NegVQA}, we leverage large language models to generate natural negations of questions from existing VQA datasets, ensuring fluency while creating challenging evaluation cases that test both linguistic and visual understanding.

We evaluate 20 state-of-the-art VLMs across seven model families and find that negation remains a major challenge. Despite their strong performance on standard VQA tasks, all models struggle significantly when faced with negated questions. For instance, Qwen2-VL-72B~\cite{qwen2vl}, the best-performing model, achieves 92.2\% accuracy on original questions but drops nearly 20 percentage points to 72.7\% on \emph{NegVQA}. Furthermore, we observe a U-shaped scaling trend, where increasing model size initially leads to worse performance on negation before eventually improving. This finding raises important questions about how VLMs process negation and how to scale up VLMs to enhance negation understanding abilities.

In summary, we propose \emph{NegVQA}, a critical diagnostic tool for evaluating negation comprehension in VLMs. Our study establishes baseline performance across major VLM families, reveals their significant shortcomings and uncovers scaling behaviors. These insights highlight the need to develop more robust and trustworthy VLMs that can accurately handle negation, a fundamental aspect of natural language understanding.

\begin{figure*}[!tb]
\centering
\begin{promptbox}
\small

\textbf{Task:}\\[0.5em]
You will be given a question collected from existing visual question answering datasets. Your task is to produce a minimally modified, negated version of the question by inserting a negation (e.g., ``not'', ``do not'', ``isn't'', etc.) in a way that:

\begin{enumerate}
\setlength{\itemsep}{0.3em}
\item \textbf{Minimal Changes:} Alters the original question as little as possible.
\item \textbf{Answer Inversion:} Causes the original correct answer to become incorrect while making one of the originally incorrect answers correct.
\item \textbf{Linguistic Accuracy:} Adheres to proper grammar and preserves the semantic intent of the question.
\end{enumerate}

\textbf{Special Case:}
\begin{enumerate}
\setlength{\itemsep}{0.3em}
\item Do not negate any background that is provided along with the question (e.g., mathematical conditions, background information, etc). Only negate the question itself (usually the last sentence).
\item If it is not possible to create a valid negation that meets these criteria, return an empty string for the negated question and set the flag \texttt{is\_negatable} to \texttt{false}.
\end{enumerate}

\textbf{Output Format:}\\[0.3em]
Your response should be an object with the following structure:

\begin{Verbatim}[fontsize=\footnotesize, commandchars=\\\{\}]
\{
  "negated_question": "<your negated question (with original background 
                       information) here, or an empty string if not negatable>",
  "is_negatable": <true/false>
\}
\end{Verbatim}

\end{promptbox}
\caption{\textbf{Detailed prompts for adding the negation using GPT-4o.}}
\label{fig:prompt}
\end{figure*}

\section{Dataset: \emph{NegVQA}}

This section details the construction and statistics of \emph{NegVQA}, our benchmark for evaluating vision language models' ability to handle negation.

\subsection{Data Curation}

We construct \emph{NegVQA} by systematically transforming questions from VMCBench~\cite{zhang2025automated}, a multi-choice visual question answering (VQA) benchmark spanning various datasets and domains, into negated versions using GPT-4o~\cite{gpt4}. Our curation process consists of two main steps.

First, we prompt GPT-4o to generate negated versions of the original questions while preserving their syntactic structure and meaning (see Figure~\ref{fig:prompt} for prompt details). For example, the question \textit{"Who wrote this book?"} is transformed into \textit{"Who did not write this book?"} We exclude questions that cannot be meaningfully negated (e.g., \textit{"Find the value of x."}), as determined by GPT-4o’s assessment of their negatability. After filtering, 7,379 out of 9,018 questions were identified as negatable and successfully transformed. To assess the accuracy of GPT-4o’s negation process, we manually verified 100 sampled negated questions and found that 97\% were correctly negated—including both the question stems and the two answer choices—confirming the high reliability of the method. Three errors are provided in Appendix Figure~\ref{fig:error}.

Second, we adjust the answer choices to reflect the negation. Each original four-choice question is reduced to a two-choice format, where we select the correct answer and randomly sample an incorrect choice, then invert their correctness. For instance, in the original question \textit{"Who wrote this book?"}, if the correct answer is \textit{"Samuelis Pufendorfii"} and an incorrect choice is \textit{"Joannem Amstelodami"}, we generate \textit{"Who did not write this book?"} where \textit{"Joannem Amstelodami"} becomes the correct answer, and \textit{"Samuelis Pufendorfii"} the incorrect one. This ensures that the negation meaningfully impacts the answer selection.

\subsection{Statistics and Examples}

\emph{NegVQA} incorporates questions from 20 widely-used VQA datasets within VMCBench, covering a broad range of vision language understanding tasks. It includes datasets for \textbf{general VQA capabilities} (VQAv2~\cite{vqav2}, OKVQA~\cite{okvqa}, MMVet~\cite{mmvet}, VizWiz~\cite{vizwiz}, A-OKVQA~\cite{aokvqa}, MMStar~\cite{mmstar}, SEEDBench~\cite{seedbench}), \textbf{reasoning tasks} (MathVision~\cite{mathvision}, GQA~\cite{gqa}, MMMU~\cite{mmmu}, RealWorldQA~\cite{realworldqa}, MathVista~\cite{mathvista}, ScienceQA~\cite{scienceqa}), \textbf{OCR-based VQA} (OCRVQA~\cite{ocrvqa}, TextVQA~\cite{textvqa}), and \textbf{document/chart comprehension} (DocVQA~\cite{docvqa}, InfoVQA~\cite{infovqa}, ChartQA~\cite{chartqa}, TableVQABench~\cite{tablevqa}, AI2D~\cite{ai2d}). The final dataset contains 7,379 questions distributed across these datasets and domains, with the detailed distribution and example questions visualized in Figure~\ref{fig:dataset}.

\emph{NegVQA} is designed to systematically test VLMs' ability to process negation in diverse visual scenarios. The dataset ensures diversity in negation forms, covering cases related to objects, attributes, logical reasoning, spatial relationships, and more. Additionally, all transformed questions have strong visual relevance, requiring models to understand both the image content and the linguistic negation to generate correct answers. \emph{NegVQA} thus serves as a comprehensive benchmark that evaluates vision language models' ability to understand negation in different visual scenarios, providing critical insights into their limitations and potential improvements.

\section{Results}

\begin{figure*}[!tb]
    \centering
    \includegraphics[width=0.495\linewidth]{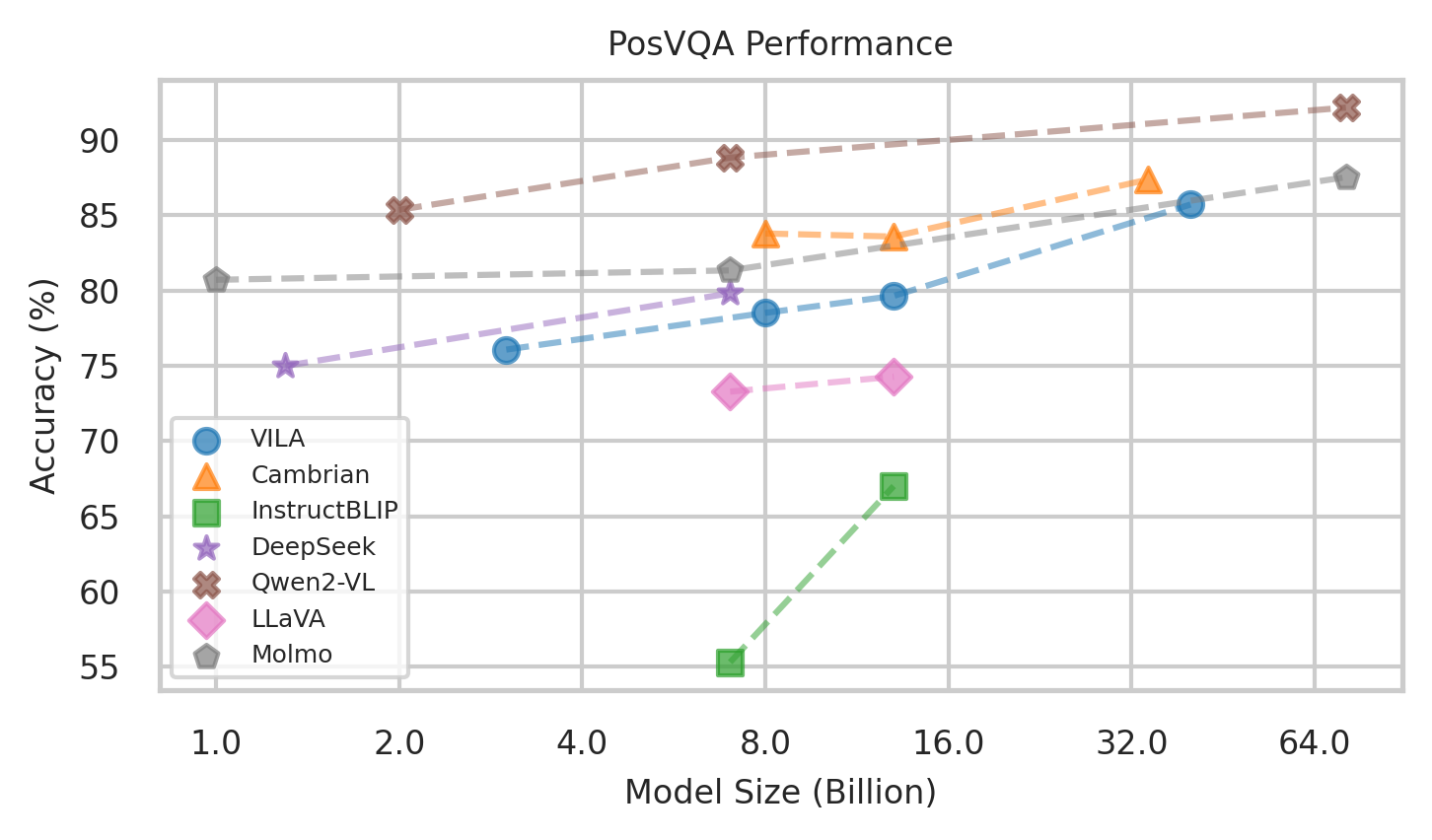}
    \includegraphics[width=0.495\linewidth]{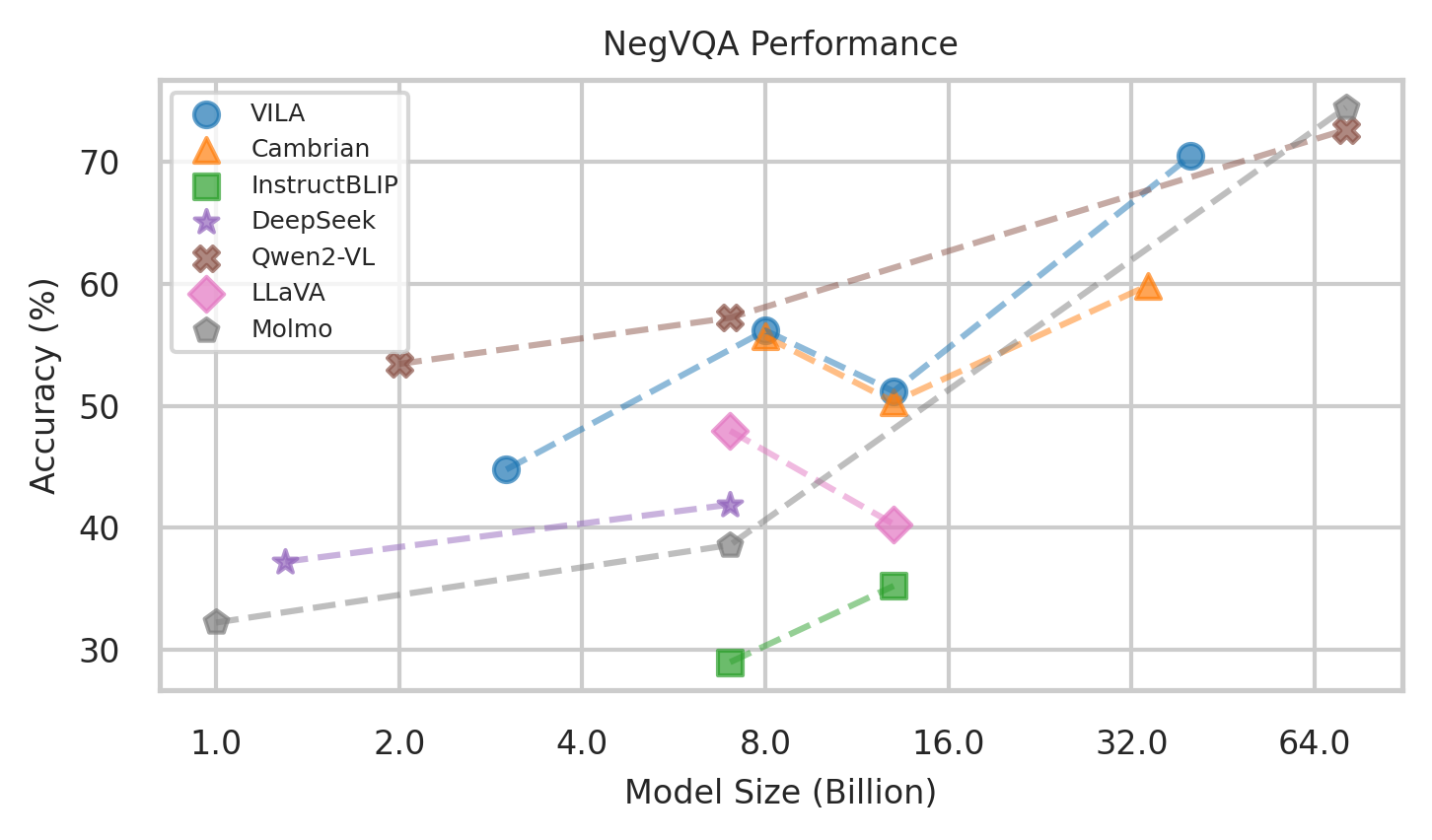}
    \includegraphics[width=0.325\linewidth]{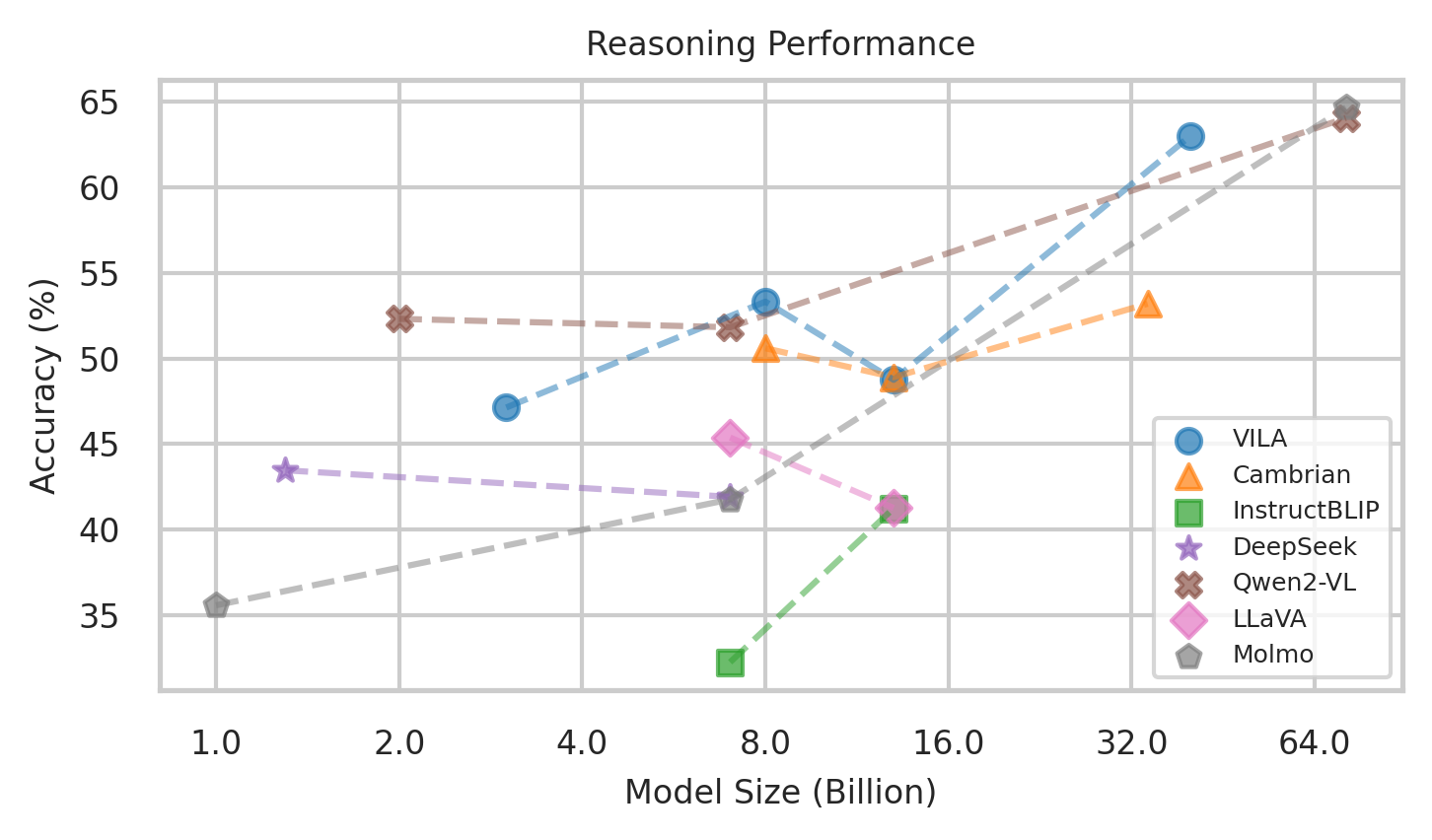}
    \includegraphics[width=0.325\linewidth]{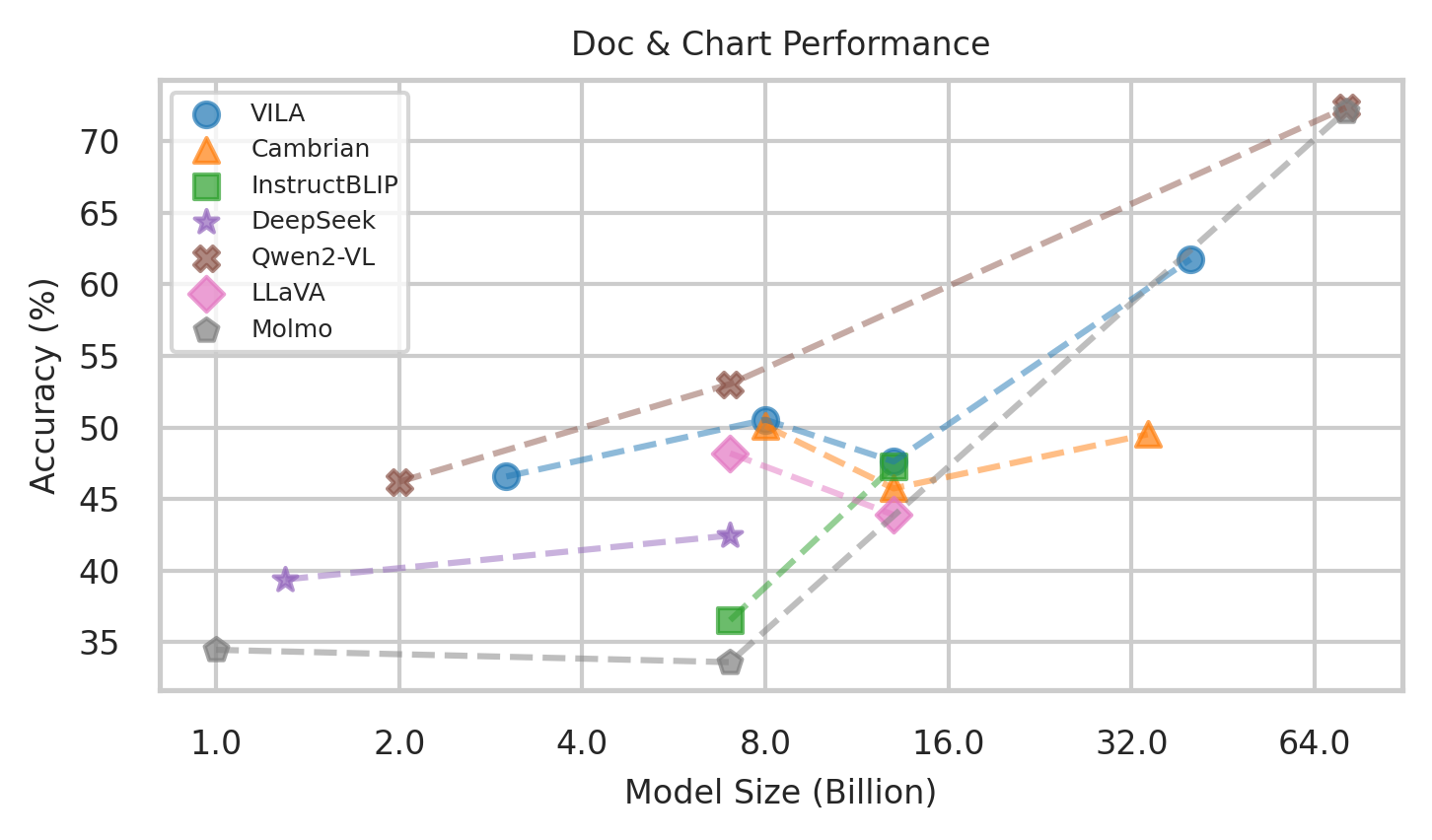}
    \includegraphics[width=0.325\linewidth]{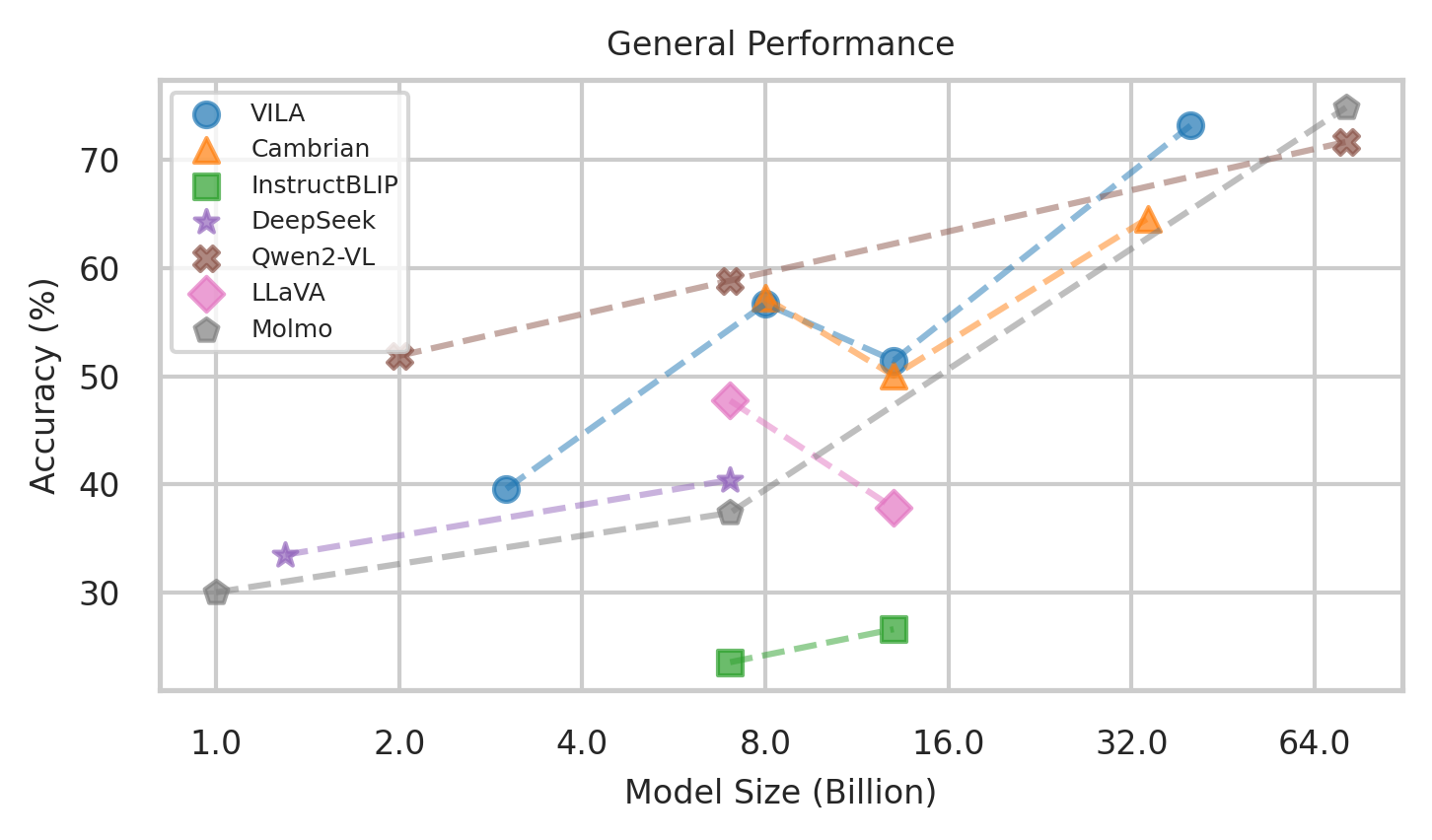}
\caption{\textbf{Model performance and scaling analysis on \emph{NegVQA} across different VLM families and task categories.} \textit{(Top left)} Performance on the original non-negated two-choice questions shows high accuracy and a positive scaling trend. \textit{(Top right)} Performance on \emph{NegVQA} (negated two-choice questions) is significantly lower, with models exhibiting a U-shaped scaling pattern—initially decreasing before improving as model size increases. \textit{(Bottom)} Category-wise breakdown of \emph{NegVQA} performance (reasoning, document/chart, general), where the U-shaped scaling effect is more pronounced in reasoning and document/chart categories.}
    \label{fig:scaling}
\end{figure*}

In this section, we describe our experimental setup and present our findings on VLM performance on \emph{NegVQA}. Our evaluation highlights two key insights: current VLMs exhibit significant difficulty in understanding negation, regardless of their size or architecture, and model scaling exhibits a U-shaped performance trend.

\subsection{Experimental Setup}

We selected 20 top-performing vision language models (VLMs) from 7 families based on the OpenVLM Leaderboard~\cite{duan2024vlmevalkit} and evaluated them on \emph{NegVQA}, including Qwen2-VL~\cite{qwen2vl}, Molmo~\cite{deitke2024molmo}, Cambrian~\cite{cambrian}, VILA~\cite{vila}, DeepSeek-VL~\cite{deepseekvl}, LLaVA1.5~\cite{llava1.5}, and InstructBLIP~\cite{instructblip}. 
These models vary in architecture, training data, and model size, providing a diverse and representative set for evaluation.
For each family, we tested multiple model sizes to analyze scaling behavior. All evaluations were conducted in a zero-shot setting using the prompt:

\begin{verbatim}
Question: <image> {question}
Options: A. {A} B. {B} C. {C} D. {D}
Answer with the option's letter from the 
given choices directly.
\end{verbatim}

The results are summarized in Figure~\ref{fig:scaling}, with detailed performance provided in Appendix Table~\ref{tab:vlm_results}.

\subsection{Findings}

\textbf{VLMs struggle with negation understanding.}
Our evaluation reveals that current VLMs consistently underperform on \emph{NegVQA} compared to the corresponding non-negated VQA tasks (which we term PosVQA). As shown in Figure~\ref{fig:scaling} (top left vs. top right), performance drops significantly across all model families when answering negated questions. The highest-performing model, Qwen2-VL-72B~\cite{qwen2vl}, achieves only 72.7\% accuracy on \emph{NegVQA}, compared to 92.2\% on non-negated questions—a gap of 19.5 percentage points. On average, model performance decreases by 29.7 points on negated questions relative to their original non-negated counterparts. This substantial decline is observed across different question types and domains, indicating a fundamental limitation in how VLMs process negation.

To contextualize model performance, we added a human baseline for the \emph{NegVQA} benchmark. We manually answered 100 questions and found that humans achieved 89\% accuracy. The remaining 11\% of errors were due to two factors: 9\% required domain-specific knowledge (mostly in subsets like MMMU and ScienceQA), and 2\% resulted from conversion errors. This 89\% human accuracy, significantly higher than the 72.7\% achieved by the state-of-the-art Qwen2-VL-72B, highlights the difficulty of negation understanding for current VLMs and the room for improvement revealed by our benchmark.

The fact that VLMs struggle with negation understanding—evidenced by the performance gaps between negated and non-negated questions, as well as between VLMs and humans—underscores a critical challenge for deploying VLMs in real-world scenarios such as robotics and other embodied environments. Appendix Table~\ref{tab:vlm_results} provides detailed numerical results. 

One potential reason VLMs struggle with negated questions is their limited exposure to negation during training. For example, in the fine-tuning data of a typical VLM like LLaVA~\cite{llava1.5}, only 1.1\% of conversations contain the word “not.” Enhancing VLMs’ ability to understand negation through training represents a promising direction for future research. One potential approach is to augment instruction-tuning datasets with carefully curated examples involving negation, thereby guiding models toward a deeper comprehension of such constructs.

\textbf{Model exhibits a U-shaped trend scaling.}
Intriguingly, a hint of a U-shaped scaling trend~\cite{wei2022inverse,zhang2023beyond} is observed: as models grow larger, their performance on \emph{NegVQA} initially declines before improving at the highest scales. This U-shaped trend is evident in model families such as Cambrian~\cite{cambrian} and VILA~\cite{vila} (Figure~\ref{fig:scaling}, top right), and is especially pronounced in reasoning and document/chart-based tasks (Figure~\ref{fig:scaling}, bottom left). Appendix Figure~\ref{fig:scaling_all} provides a detailed breakdown of performance across individual datasets.

Conceptually, this U-shaped trend can be understood as the composition of two underlying capabilities: original question answering, which tends to improve steadily with model scale, and negation understanding, which follows a tanh-like activation curve. Smaller models with limited reasoning ability often treat negated questions as if they were non-negated, ignoring the negation and selecting answers accordingly. As models scale up, their performance on non-negated questions improves, but their misunderstanding of negation becomes more detrimental, leading to a dip in performance on negated questions. Only when models reach a sufficient level of sophistication to handle negation properly does their performance on negated questions recover, completing the U-shaped trajectory.

Overall, these results underscore the persistent challenges VLMs face in handling negation and highlight the intriguing scaling behavior of VLMs.

\section{Related Work}

\textbf{Vision language models (VLMs).} VLMs enable multimodal understanding by modeling \( p(y_t | y_{<t}, x) \) in an auto-regressive manner, where \( y_i \) represents text tokens and \( x \) represents visual input. Modern VLMs typically comprise three key components: a visual encoder (often CLIP~\cite{radford2021learning}), a language model, and a linear or MLP projector connecting them. Notable examples include proprietary models such as GPT-4o~\cite{gpt4} and Claude~\cite{claude}, as well as open-source models like LLaVA~\cite{llava1.5} and BLIP~\cite{li2023blip}. These models are generally trained on image-text pairs and instruction-tuning datasets, leveraging pre-trained vision and language components. While they exhibit strong performance on various image understanding tasks~\cite{llava1.5,deitke2024molmo,qwen2vl} and have been applied in embodied AI and robotics~\cite{driess2023palm,brohan2023rt,kim2024openvla}, their ability to handle negation remains largely unexplored.

\textbf{Negation understanding.} Negation plays a fundamental role in language comprehension~\cite{ackrill1975categories}. Most prior research has focused on evaluating language models' ability to understand negation~\cite{hossain-etal-2020-analysis,fancellu-webber-2015-translating-negation,kassner-schutze-2020-negated,zhang2023beyond}. More recently, studies have begun assessing CLIP~\cite{radford2021learning}'s understanding of negation~\cite{alhamoud2025vision,singh2024learn,quantmeyer2024and}. However, to the best of our knowledge, no prior work has systematically evaluated negation comprehension in generative VLMs. In this work, we introduce \emph{NegVQA}, the first benchmark designed to assess VLMs' ability to handle negation. Given the increasing deployment of VLMs in real-world embodied AI systems, understanding their limitations in processing negation is crucial, as failures in user intent interpretation could lead to unintended and risky scenarios.

\textbf{Scaling trends.} Scaling up models has been a dominant approach in advancing foundation models. However, most scaling studies have focused on language models~\cite{kaplan2020scaling,brown2020language,ruan2024observational}. While many tasks benefit from scaling, some exhibit inverse scaling~\cite{lin-etal-2022-truthfulqa,mckenzieinverse} or U-shaped scaling~\cite{wei2022inverse,zhang2023beyond}. In this work, we analyze scaling effects in vision language models on the negation task and reveal a similar U-shaped scaling pattern.

\section{Conclusion}

In this work, we present \emph{NegVQA}, a benchmark designed to evaluate vision language models’ ability to comprehend negation. Our analysis of 20 VLMs highlights their significant limitations in handling negation and uncovers a U-shaped scaling pattern in performance. We envision \emph{NegVQA} as a valuable resource for advancing linguistically competent, safe, and trustworthy vision language models.

\paragraph{Acknowledgments.}

S.Y. is a Chan Zuckerberg Biohub — San Francisco Investigator.

\newpage
\section*{Limitations}
Our study has three limitations: First, while our multiple-choice format enables controlled experimentation and easy evaluation metrics, it may not fully capture how VLMs handle negation in more open-ended or real-world scenarios where models cannot rely on predefined answer choices. Second, we focus exclusively on zero-shot evaluation, due to current VLMs' architectural constraint of accepting only single image inputs, leaving unexplored how few-shot prompting might affect negation understanding and performance scaling. Finally, this work primarily investigates how vision-language models (VLMs) handle negation. Enhancing their ability to understand and process negation during training is a promising direction for future research. One potential approach is to augment instruction-tuning datasets with carefully curated examples involving negation, thereby guiding models toward a deeper comprehension of such constructs. Despite these limitations, our work provides the first comprehensive analysis of how VLMs process negation, uncovering both their current limitations and a U-shaped scaling pattern. The \emph{NegVQA} benchmark establishes a foundation for systematically evaluating and improving how future vision language models handle this fundamental linguistic operation.

\bibliography{custom}

\newpage
\appendix

\begin{table*}[!tb]
\centering
\rowcolors{2}{white}{light-light-gray}
\setlength\tabcolsep{1pt}
\renewcommand{\arraystretch}{1.2}
\small
\begin{tabular}{lp{0.04\linewidth}cccccp{0.04\linewidth}ccccc}
\toprule
 & & \multicolumn{5}{c}{\textbf{Original Non-negated Questions}} & & \multicolumn{5}{c}{\textbf{Negated Questions (\emph{NegVQA})}} \\
\textbf{Model} & & \textbf{General} & \textbf{Reason} & \textbf{OCR} & \textbf{Doc\&Cht} & \textbf{Average} & & \textbf{General} & \textbf{Reason} & \textbf{OCR} & \textbf{Doc\&Cht} & \textbf{Average} \\
\midrule
Cambrian-8B && 87.6\hspace{0.5mm}\scriptsize{$_{86.3}^{88.9}$} & 74.0\hspace{0.5mm}\scriptsize{$_{72.1}^{75.9}$} & 93.4\hspace{0.5mm}\scriptsize{$_{91.5}^{95.3}$} & 80.9\hspace{0.5mm}\scriptsize{$_{79.2}^{82.6}$} & 83.8\hspace{0.5mm}\scriptsize{$_{83.0}^{84.6}$} & & 57.2\hspace{0.5mm}\scriptsize{$_{55.3}^{59.1}$} & 50.6\hspace{0.5mm}\scriptsize{$_{48.4}^{52.8}$} & 71.8\hspace{0.5mm}\scriptsize{$_{68.3}^{75.3}$} & 50.1\hspace{0.5mm}\scriptsize{$_{48.0}^{52.2}$} & 55.7\hspace{0.5mm}\scriptsize{$_{54.6}^{56.8}$} \\
Cambrian-13B && 87.7\hspace{0.5mm}\scriptsize{$_{86.4}^{89.0}$} & 73.5\hspace{0.5mm}\scriptsize{$_{71.6}^{75.4}$} & 95.9\hspace{0.5mm}\scriptsize{$_{94.4}^{97.4}$} & 80.5\hspace{0.5mm}\scriptsize{$_{78.8}^{82.2}$} & 83.6\hspace{0.5mm}\scriptsize{$_{82.8}^{84.4}$} & & 50.1\hspace{0.5mm}\scriptsize{$_{48.2}^{52.0}$} & 48.9\hspace{0.5mm}\scriptsize{$_{46.7}^{51.1}$} & 69.2\hspace{0.5mm}\scriptsize{$_{65.6}^{72.8}$} & 45.7\hspace{0.5mm}\scriptsize{$_{43.6}^{47.8}$} & 50.3\hspace{0.5mm}\scriptsize{$_{49.2}^{51.4}$} \\
Cambrian-34B && 90.0\hspace{0.5mm}\scriptsize{$_{88.9}^{91.1}$} & 80.3\hspace{0.5mm}\scriptsize{$_{78.6}^{82.0}$} & 96.6\hspace{0.5mm}\scriptsize{$_{95.2}^{98.0}$} & 85.0\hspace{0.5mm}\scriptsize{$_{83.5}^{86.5}$} & 87.4\hspace{0.5mm}\scriptsize{$_{86.6}^{88.2}$} & & 64.6\hspace{0.5mm}\scriptsize{$_{62.8}^{66.4}$} & 53.2\hspace{0.5mm}\scriptsize{$_{51.0}^{55.4}$} & 81.5\hspace{0.5mm}\scriptsize{$_{78.5}^{84.5}$} & 49.5\hspace{0.5mm}\scriptsize{$_{47.4}^{51.6}$} & 59.9\hspace{0.5mm}\scriptsize{$_{58.8}^{61.0}$} \\
\midrule
InstructBLIP-7B && 58.5\hspace{0.5mm}\scriptsize{$_{56.6}^{60.4}$} & 53.9\hspace{0.5mm}\scriptsize{$_{51.7}^{56.1}$} & 70.0\hspace{0.5mm}\scriptsize{$_{66.4}^{73.6}$} & 48.4\hspace{0.5mm}\scriptsize{$_{46.3}^{50.5}$} & 55.3\hspace{0.5mm}\scriptsize{$_{54.2}^{56.4}$} & & 23.6\hspace{0.5mm}\scriptsize{$_{22.0}^{25.2}$} & 32.2\hspace{0.5mm}\scriptsize{$_{30.1}^{34.3}$} & 20.7\hspace{0.5mm}\scriptsize{$_{17.5}^{23.9}$} & 36.5\hspace{0.5mm}\scriptsize{$_{34.4}^{38.6}$} & 28.9\hspace{0.5mm}\scriptsize{$_{27.9}^{29.9}$} \\
InstructBLIP-13B && 75.8\hspace{0.5mm}\scriptsize{$_{74.2}^{77.4}$} & 62.5\hspace{0.5mm}\scriptsize{$_{60.4}^{64.6}$} & 68.1\hspace{0.5mm}\scriptsize{$_{64.5}^{71.7}$} & 53.9\hspace{0.5mm}\scriptsize{$_{51.8}^{56.0}$} & 67.0\hspace{0.5mm}\scriptsize{$_{65.9}^{68.1}$} & & 26.6\hspace{0.5mm}\scriptsize{$_{24.9}^{28.3}$} & 41.2\hspace{0.5mm}\scriptsize{$_{39.0}^{43.4}$} & 20.3\hspace{0.5mm}\scriptsize{$_{17.2}^{23.4}$} & 47.3\hspace{0.5mm}\scriptsize{$_{45.2}^{49.4}$} & 35.2\hspace{0.5mm}\scriptsize{$_{34.1}^{36.3}$} \\
\midrule
DeepSeek-VL-1.3B && 81.5\hspace{0.5mm}\scriptsize{$_{80.0}^{83.0}$} & 66.6\hspace{0.5mm}\scriptsize{$_{64.5}^{68.7}$} & 88.6\hspace{0.5mm}\scriptsize{$_{86.1}^{91.1}$} & 65.9\hspace{0.5mm}\scriptsize{$_{63.9}^{67.9}$} & 75.0\hspace{0.5mm}\scriptsize{$_{74.0}^{76.0}$} & & 33.5\hspace{0.5mm}\scriptsize{$_{31.7}^{35.3}$} & 43.5\hspace{0.5mm}\scriptsize{$_{41.3}^{45.7}$} & 34.6\hspace{0.5mm}\scriptsize{$_{30.9}^{38.3}$} & 39.4\hspace{0.5mm}\scriptsize{$_{37.3}^{41.5}$} & 37.2\hspace{0.5mm}\scriptsize{$_{36.1}^{38.3}$} \\
DeepSeek-VL-7B && 84.7\hspace{0.5mm}\scriptsize{$_{83.3}^{86.1}$} & 71.2\hspace{0.5mm}\scriptsize{$_{69.2}^{73.2}$} & 91.3\hspace{0.5mm}\scriptsize{$_{89.1}^{93.5}$} & 73.1\hspace{0.5mm}\scriptsize{$_{71.2}^{75.0}$} & 79.8\hspace{0.5mm}\scriptsize{$_{78.9}^{80.7}$} & & 40.4\hspace{0.5mm}\scriptsize{$_{38.5}^{42.3}$} & 41.9\hspace{0.5mm}\scriptsize{$_{39.7}^{44.1}$} & 53.7\hspace{0.5mm}\scriptsize{$_{49.8}^{57.6}$} & 42.4\hspace{0.5mm}\scriptsize{$_{40.3}^{44.5}$} & 41.9\hspace{0.5mm}\scriptsize{$_{40.8}^{43.0}$} \\
\midrule
LLaVA-1.5-7B && 81.0\hspace{0.5mm}\scriptsize{$_{79.5}^{82.5}$} & 67.7\hspace{0.5mm}\scriptsize{$_{65.6}^{69.8}$} & 85.5\hspace{0.5mm}\scriptsize{$_{82.7}^{88.3}$} & 61.1\hspace{0.5mm}\scriptsize{$_{59.0}^{63.2}$} & 73.3\hspace{0.5mm}\scriptsize{$_{72.3}^{74.3}$} & & 47.7\hspace{0.5mm}\scriptsize{$_{45.8}^{49.6}$} & 45.4\hspace{0.5mm}\scriptsize{$_{43.2}^{47.6}$} & 49.7\hspace{0.5mm}\scriptsize{$_{45.8}^{53.6}$} & 48.2\hspace{0.5mm}\scriptsize{$_{46.1}^{50.3}$} & 47.9\hspace{0.5mm}\scriptsize{$_{46.8}^{49.0}$} \\
LLaVA-1.5-13B && 82.8\hspace{0.5mm}\scriptsize{$_{81.4}^{84.2}$} & 66.5\hspace{0.5mm}\scriptsize{$_{64.4}^{68.6}$} & 86.4\hspace{0.5mm}\scriptsize{$_{83.7}^{89.1}$} & 62.3\hspace{0.5mm}\scriptsize{$_{60.2}^{64.4}$} & 74.3\hspace{0.5mm}\scriptsize{$_{73.3}^{75.3}$} & & 37.8\hspace{0.5mm}\scriptsize{$_{36.0}^{39.6}$} & 41.2\hspace{0.5mm}\scriptsize{$_{39.0}^{43.4}$} & 40.4\hspace{0.5mm}\scriptsize{$_{36.6}^{44.2}$} & 43.9\hspace{0.5mm}\scriptsize{$_{41.8}^{46.0}$} & 40.3\hspace{0.5mm}\scriptsize{$_{39.2}^{41.4}$} \\
\midrule
Molmo-1B && 83.6\hspace{0.5mm}\scriptsize{$_{82.2}^{85.0}$} & 71.7\hspace{0.5mm}\scriptsize{$_{69.7}^{73.7}$} & 92.0\hspace{0.5mm}\scriptsize{$_{89.9}^{94.1}$} & 77.7\hspace{0.5mm}\scriptsize{$_{75.9}^{79.5}$} & 80.7\hspace{0.5mm}\scriptsize{$_{79.8}^{81.6}$} & & 30.0\hspace{0.5mm}\scriptsize{$_{28.3}^{31.7}$} & 35.6\hspace{0.5mm}\scriptsize{$_{33.5}^{37.7}$} & 30.4\hspace{0.5mm}\scriptsize{$_{26.8}^{34.0}$} & 34.5\hspace{0.5mm}\scriptsize{$_{32.5}^{36.5}$} & 32.2\hspace{0.5mm}\scriptsize{$_{31.1}^{33.3}$} \\
Molmo-7B-O && 83.1\hspace{0.5mm}\scriptsize{$_{81.7}^{84.5}$} & 69.9\hspace{0.5mm}\scriptsize{$_{67.9}^{71.9}$} & 91.2\hspace{0.5mm}\scriptsize{$_{89.0}^{93.4}$} & 81.4\hspace{0.5mm}\scriptsize{$_{79.7}^{83.1}$} & 81.3\hspace{0.5mm}\scriptsize{$_{80.4}^{82.2}$} & & 37.4\hspace{0.5mm}\scriptsize{$_{35.6}^{39.2}$} & 41.7\hspace{0.5mm}\scriptsize{$_{39.5}^{43.9}$} & 49.4\hspace{0.5mm}\scriptsize{$_{45.5}^{53.3}$} & 33.6\hspace{0.5mm}\scriptsize{$_{31.6}^{35.6}$} & 38.6\hspace{0.5mm}\scriptsize{$_{37.5}^{39.7}$} \\
Molmo-7B-D && 85.6\hspace{0.5mm}\scriptsize{$_{84.3}^{86.9}$} & 67.8\hspace{0.5mm}\scriptsize{$_{65.7}^{69.9}$} & 94.8\hspace{0.5mm}\scriptsize{$_{93.1}^{96.5}$} & 84.3\hspace{0.5mm}\scriptsize{$_{82.7}^{85.9}$} & 83.0\hspace{0.5mm}\scriptsize{$_{82.1}^{83.9}$} & & 55.9\hspace{0.5mm}\scriptsize{$_{54.0}^{57.8}$} & 48.6\hspace{0.5mm}\scriptsize{$_{46.4}^{50.8}$} & 75.3\hspace{0.5mm}\scriptsize{$_{71.9}^{78.7}$} & 49.7\hspace{0.5mm}\scriptsize{$_{47.6}^{51.8}$} & 55.3\hspace{0.5mm}\scriptsize{$_{54.2}^{56.4}$} \\
Molmo-72B && 89.4\hspace{0.5mm}\scriptsize{$_{88.2}^{90.6}$} & 78.2\hspace{0.5mm}\scriptsize{$_{76.4}^{80.0}$} & 96.7\hspace{0.5mm}\scriptsize{$_{95.3}^{98.1}$} & 89.0\hspace{0.5mm}\scriptsize{$_{87.7}^{90.3}$} & 87.5\hspace{0.5mm}\scriptsize{$_{86.7}^{88.3}$} & & 74.8\hspace{0.5mm}\scriptsize{$_{73.1}^{76.5}$} & 64.7\hspace{0.5mm}\scriptsize{$_{62.6}^{66.8}$} & 93.9\hspace{0.5mm}\scriptsize{$_{92.0}^{95.8}$} & 72.1\hspace{0.5mm}\scriptsize{$_{70.2}^{74.0}$} & 74.5\hspace{0.5mm}\scriptsize{$_{73.5}^{75.5}$} \\
\midrule
Qwen2-VL-2B && 88.6\hspace{0.5mm}\scriptsize{$_{87.4}^{89.8}$} & 74.7\hspace{0.5mm}\scriptsize{$_{72.8}^{76.6}$} & 96.1\hspace{0.5mm}\scriptsize{$_{94.6}^{97.6}$} & 84.8\hspace{0.5mm}\scriptsize{$_{83.3}^{86.3}$} & 85.4\hspace{0.5mm}\scriptsize{$_{84.6}^{86.2}$} & & 51.9\hspace{0.5mm}\scriptsize{$_{50.0}^{53.8}$} & 52.3\hspace{0.5mm}\scriptsize{$_{50.1}^{54.5}$} & 78.0\hspace{0.5mm}\scriptsize{$_{74.8}^{81.2}$} & 46.2\hspace{0.5mm}\scriptsize{$_{44.1}^{48.3}$} & 53.4\hspace{0.5mm}\scriptsize{$_{52.3}^{54.5}$} \\
Qwen2-VL-7B && 91.3\hspace{0.5mm}\scriptsize{$_{90.2}^{92.4}$} & 79.8\hspace{0.5mm}\scriptsize{$_{78.0}^{81.6}$} & 97.2\hspace{0.5mm}\scriptsize{$_{95.9}^{98.5}$} & 89.4\hspace{0.5mm}\scriptsize{$_{88.1}^{90.7}$} & 88.8\hspace{0.5mm}\scriptsize{$_{88.1}^{89.5}$} & & 58.8\hspace{0.5mm}\scriptsize{$_{56.9}^{60.7}$} & 51.8\hspace{0.5mm}\scriptsize{$_{49.6}^{54.0}$} & 82.0\hspace{0.5mm}\scriptsize{$_{79.0}^{85.0}$} & 53.0\hspace{0.5mm}\scriptsize{$_{50.9}^{55.1}$} & 57.2\hspace{0.5mm}\scriptsize{$_{56.1}^{58.3}$} \\
Qwen2-VL-72B && 93.6\hspace{0.5mm}\scriptsize{$_{92.7}^{94.5}$} & 83.4\hspace{0.5mm}\scriptsize{$_{81.8}^{85.0}$} & 99.0\hspace{0.5mm}\scriptsize{$_{98.2}^{99.8}$} & 94.8\hspace{0.5mm}\scriptsize{$_{93.9}^{95.7}$} & 92.2\hspace{0.5mm}\scriptsize{$_{91.6}^{92.8}$} & & 71.7\hspace{0.5mm}\scriptsize{$_{70.0}^{73.4}$} & 64.1\hspace{0.5mm}\scriptsize{$_{62.0}^{66.2}$} & 91.8\hspace{0.5mm}\scriptsize{$_{89.7}^{93.9}$} & 72.4\hspace{0.5mm}\scriptsize{$_{70.5}^{74.3}$} & 72.7\hspace{0.5mm}\scriptsize{$_{71.7}^{73.7}$} \\
\midrule
VILA1.5-3B && 83.9\hspace{0.5mm}\scriptsize{$_{82.5}^{85.3}$} & 68.0\hspace{0.5mm}\scriptsize{$_{66.0}^{70.0}$} & 88.2\hspace{0.5mm}\scriptsize{$_{85.7}^{90.7}$} & 66.0\hspace{0.5mm}\scriptsize{$_{64.0}^{68.0}$} & 76.1\hspace{0.5mm}\scriptsize{$_{75.1}^{77.1}$} & & 39.6\hspace{0.5mm}\scriptsize{$_{37.7}^{41.5}$} & 47.1\hspace{0.5mm}\scriptsize{$_{44.9}^{49.3}$} & 51.9\hspace{0.5mm}\scriptsize{$_{48.0}^{55.8}$} & 46.6\hspace{0.5mm}\scriptsize{$_{44.5}^{48.7}$} & 44.8\hspace{0.5mm}\scriptsize{$_{43.7}^{45.9}$} \\
VILA1.5-8B && 85.3\hspace{0.5mm}\scriptsize{$_{84.0}^{86.6}$} & 71.2\hspace{0.5mm}\scriptsize{$_{69.2}^{73.2}$} & 91.0\hspace{0.5mm}\scriptsize{$_{88.8}^{93.2}$} & 69.4\hspace{0.5mm}\scriptsize{$_{67.4}^{71.4}$} & 78.5\hspace{0.5mm}\scriptsize{$_{77.6}^{79.4}$} & & 56.7\hspace{0.5mm}\scriptsize{$_{54.8}^{58.6}$} & 53.3\hspace{0.5mm}\scriptsize{$_{51.1}^{55.5}$} & 68.4\hspace{0.5mm}\scriptsize{$_{64.8}^{72.0}$} & 50.5\hspace{0.5mm}\scriptsize{$_{48.4}^{52.6}$} & 56.2\hspace{0.5mm}\scriptsize{$_{55.1}^{57.3}$} \\
VILA1.5-13B && 85.7\hspace{0.5mm}\scriptsize{$_{84.4}^{87.0}$} & 73.7\hspace{0.5mm}\scriptsize{$_{71.8}^{75.6}$} & 91.6\hspace{0.5mm}\scriptsize{$_{89.4}^{93.8}$} & 70.3\hspace{0.5mm}\scriptsize{$_{68.4}^{72.2}$} & 79.6\hspace{0.5mm}\scriptsize{$_{78.7}^{80.5}$} & & 51.4\hspace{0.5mm}\scriptsize{$_{49.5}^{53.3}$} & 48.7\hspace{0.5mm}\scriptsize{$_{46.5}^{50.9}$} & 62.5\hspace{0.5mm}\scriptsize{$_{58.7}^{66.3}$} & 47.6\hspace{0.5mm}\scriptsize{$_{45.5}^{49.7}$} & 51.2\hspace{0.5mm}\scriptsize{$_{50.1}^{52.3}$} \\
VILA1.5-40B && 89.4\hspace{0.5mm}\scriptsize{$_{88.2}^{90.6}$} & 78.6\hspace{0.5mm}\scriptsize{$_{76.8}^{80.4}$} & 96.3\hspace{0.5mm}\scriptsize{$_{94.8}^{97.8}$} & 81.5\hspace{0.5mm}\scriptsize{$_{79.8}^{83.2}$} & 85.7\hspace{0.5mm}\scriptsize{$_{84.9}^{86.5}$} & & 73.2\hspace{0.5mm}\scriptsize{$_{71.5}^{74.9}$} & 63.0\hspace{0.5mm}\scriptsize{$_{60.9}^{65.1}$} & 90.3\hspace{0.5mm}\scriptsize{$_{88.0}^{92.6}$} & 61.8\hspace{0.5mm}\scriptsize{$_{59.7}^{63.9}$} & 70.5\hspace{0.5mm}\scriptsize{$_{69.5}^{71.5}$} \\
\bottomrule
\end{tabular}
\vspace{-1em}
\caption{\textbf{Performance of 20 vision language models from 7 families on \textit{\emph{NegVQA}} and the original non-negated dataset.} Each reported accuracy is accompanied by a 95\% binomial confidence interval, with the lower bound shown as a subscript and the upper bound as a superscript. }
\vspace{-1em}
\label{tab:vlm_results}
\end{table*}

\begin{figure*}[!tb]
\centering
\begin{errorbox}
\scriptsize

\begin{examplebox}{Example 1: Improper Negation}
\textcolor{originalcolor}{\textbf{Original Question:}} how many total singles does he have?\\[0.2em]
\textcolor{negatedcolor}{\textbf{Negated Question:}} how many total singles does he \textbf{not have}?
\end{examplebox}

\begin{examplebox}{Example 2: Condition Negation Error}
\textcolor{originalcolor}{\textbf{Original Question:}} As shown in the figure, points A, B, and C are three points on O, and the straight line CD and O are tangent to point C. If DCB = 40.0, then the degree of CAB is ()\\[0.2em]
\textcolor{negatedcolor}{\textbf{Negated Question:}} As shown in the figure, points A, B, and C are three points on O, and the straight line CD and O are \textbf{not tangent} to point C. If DCB = 40.0, then the degree of CAB is ()
\end{examplebox}

\begin{examplebox}{Example 3: Condition Negation Error}
\textcolor{originalcolor}{\textbf{Original Question:}} If cricket was removed from the food web, there would be\\[0.2em]
\textcolor{negatedcolor}{\textbf{Negated Question:}} If cricket was \textbf{not removed} from the food web, there would be
\end{examplebox}

\end{errorbox}
\caption{\textbf{Errors in negated questions generated by GPT-4o.} The first question cannot be negated, while the second and third questions are negated in the condition, whereas the negation should apply to the main question.}
\label{fig:error}
\end{figure*}

\begin{figure*}
    \centering
\includegraphics[width=0.245\textwidth]{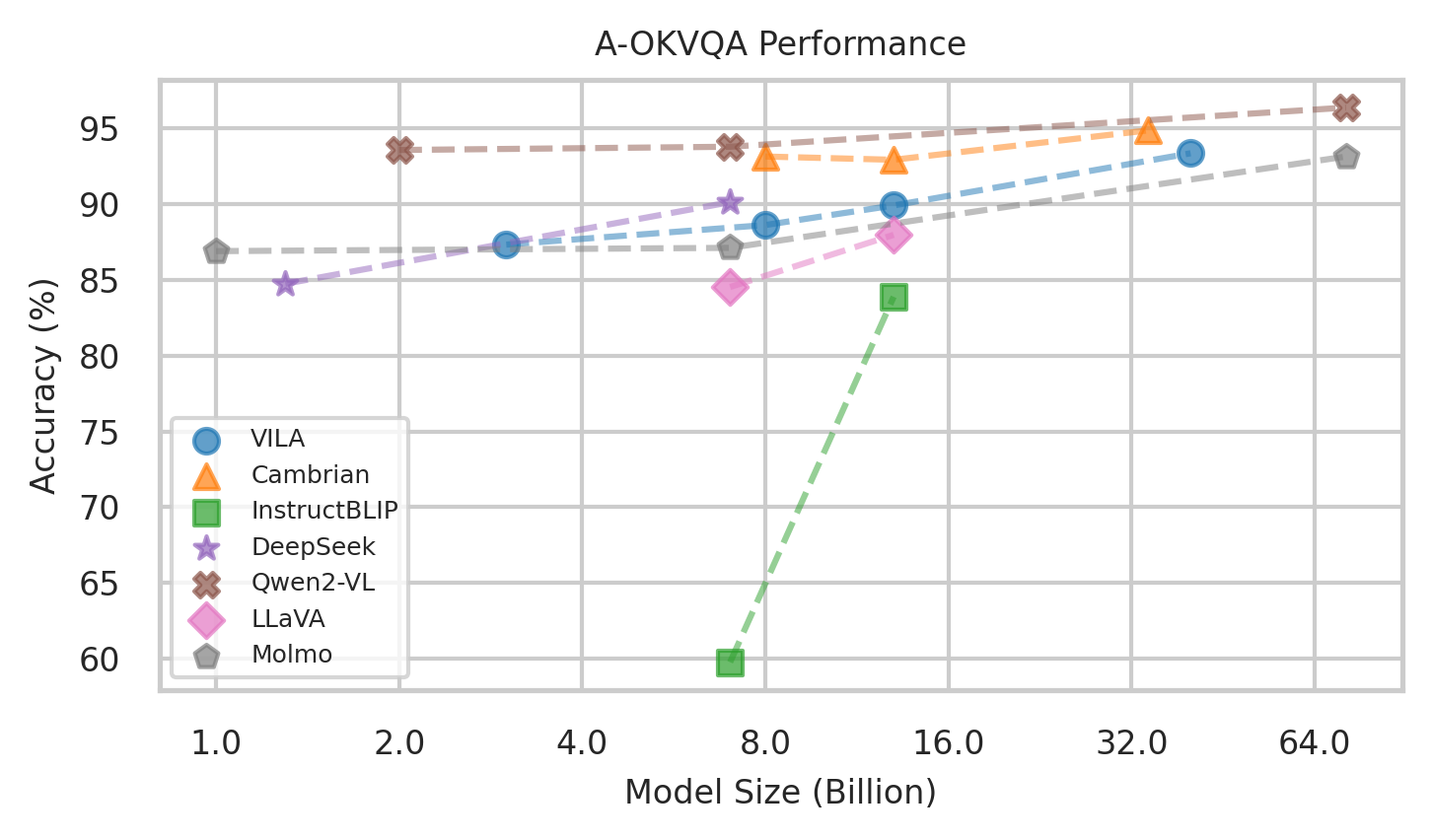}
\includegraphics[width=0.245\textwidth]{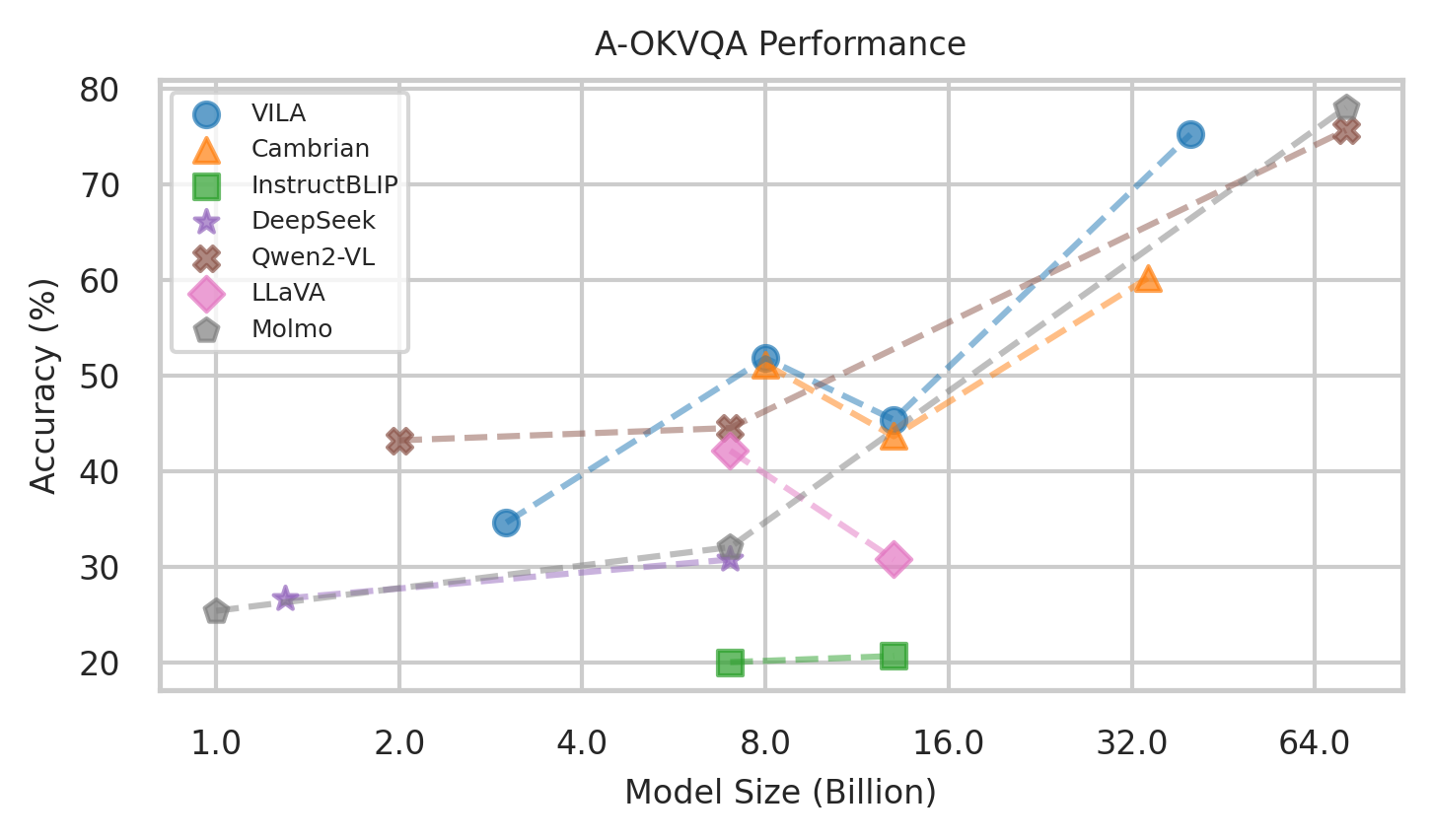}
\includegraphics[width=0.245\textwidth]{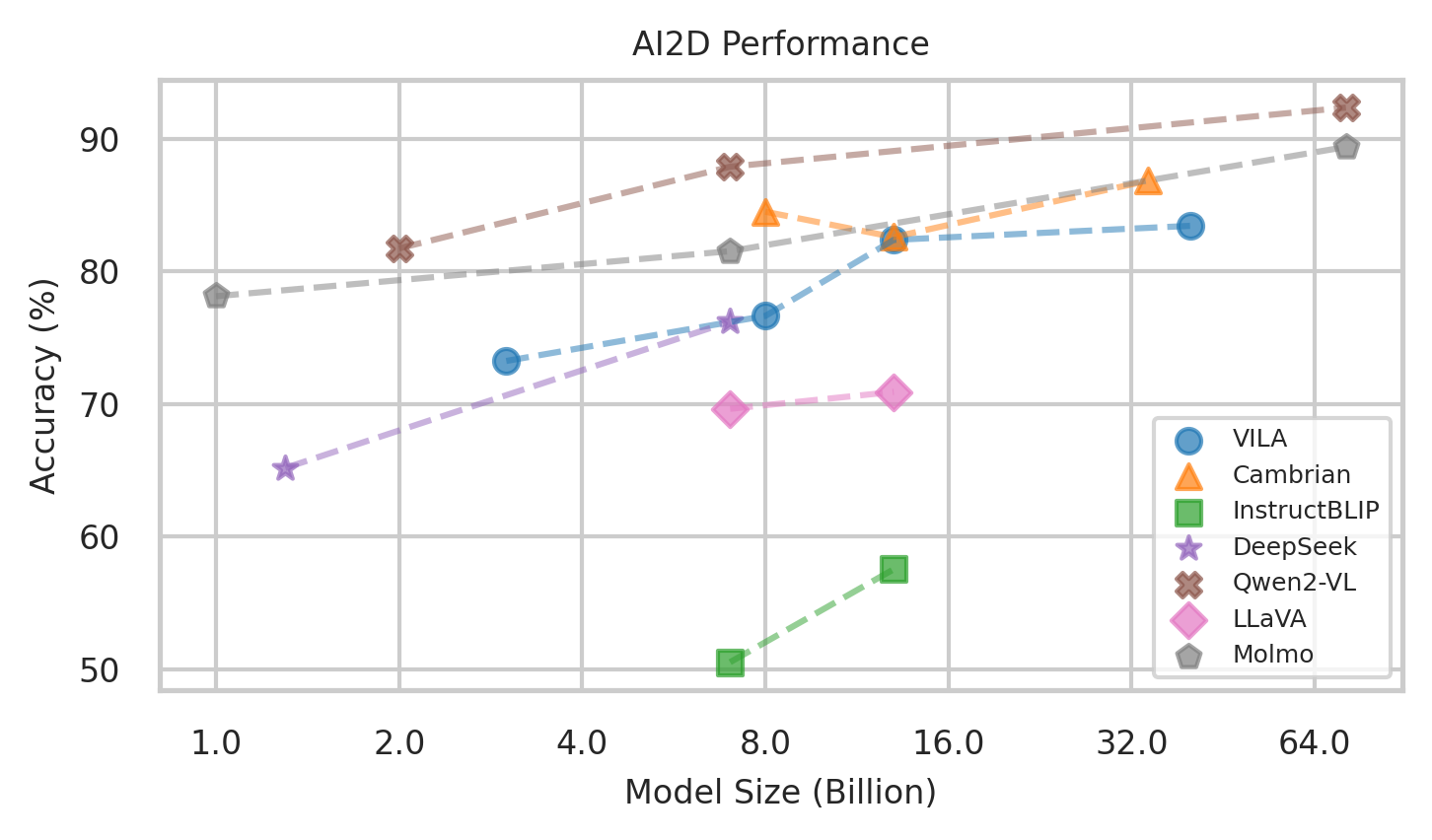}
\includegraphics[width=0.245\textwidth]{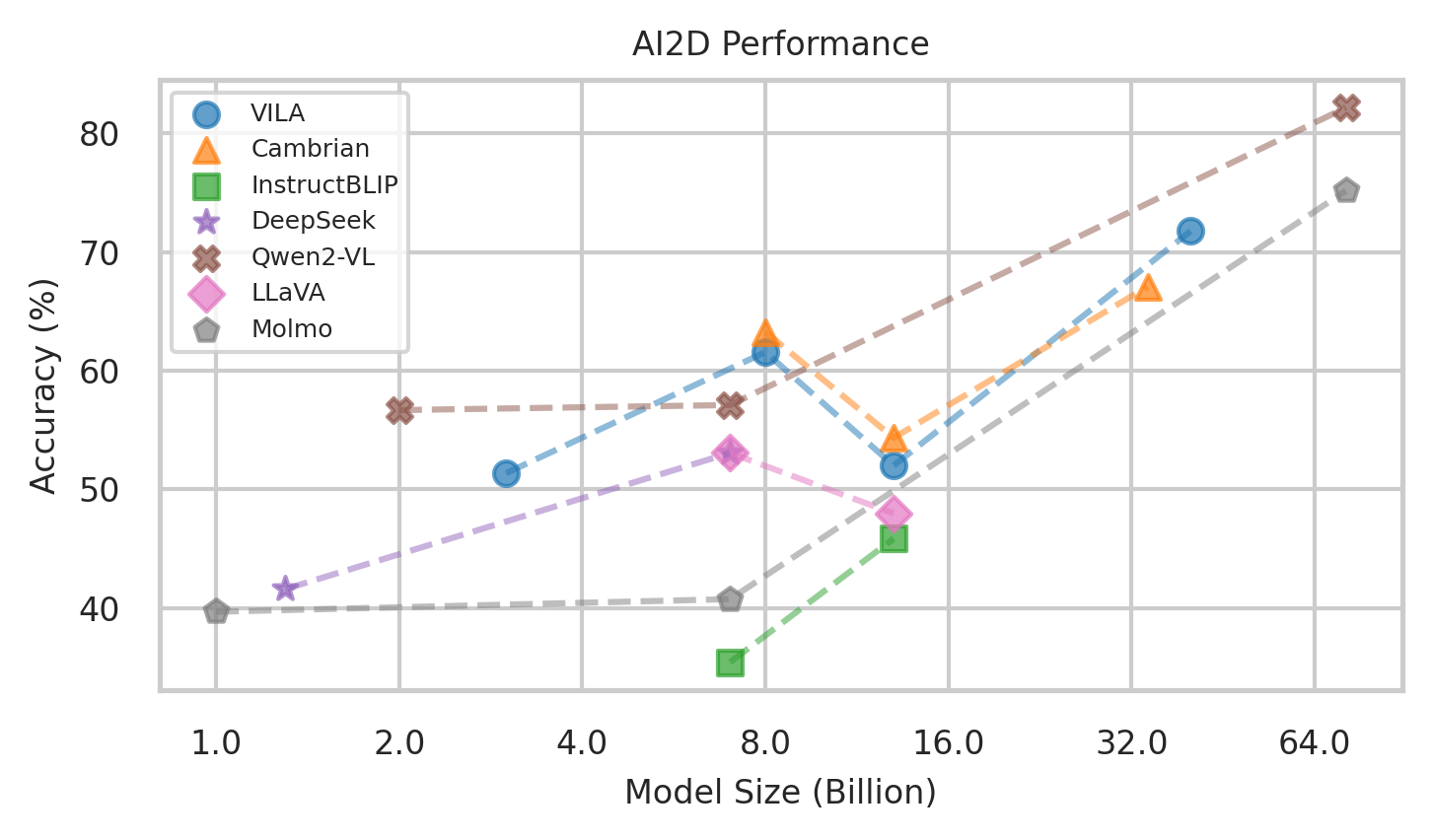}
\includegraphics[width=0.245\textwidth]{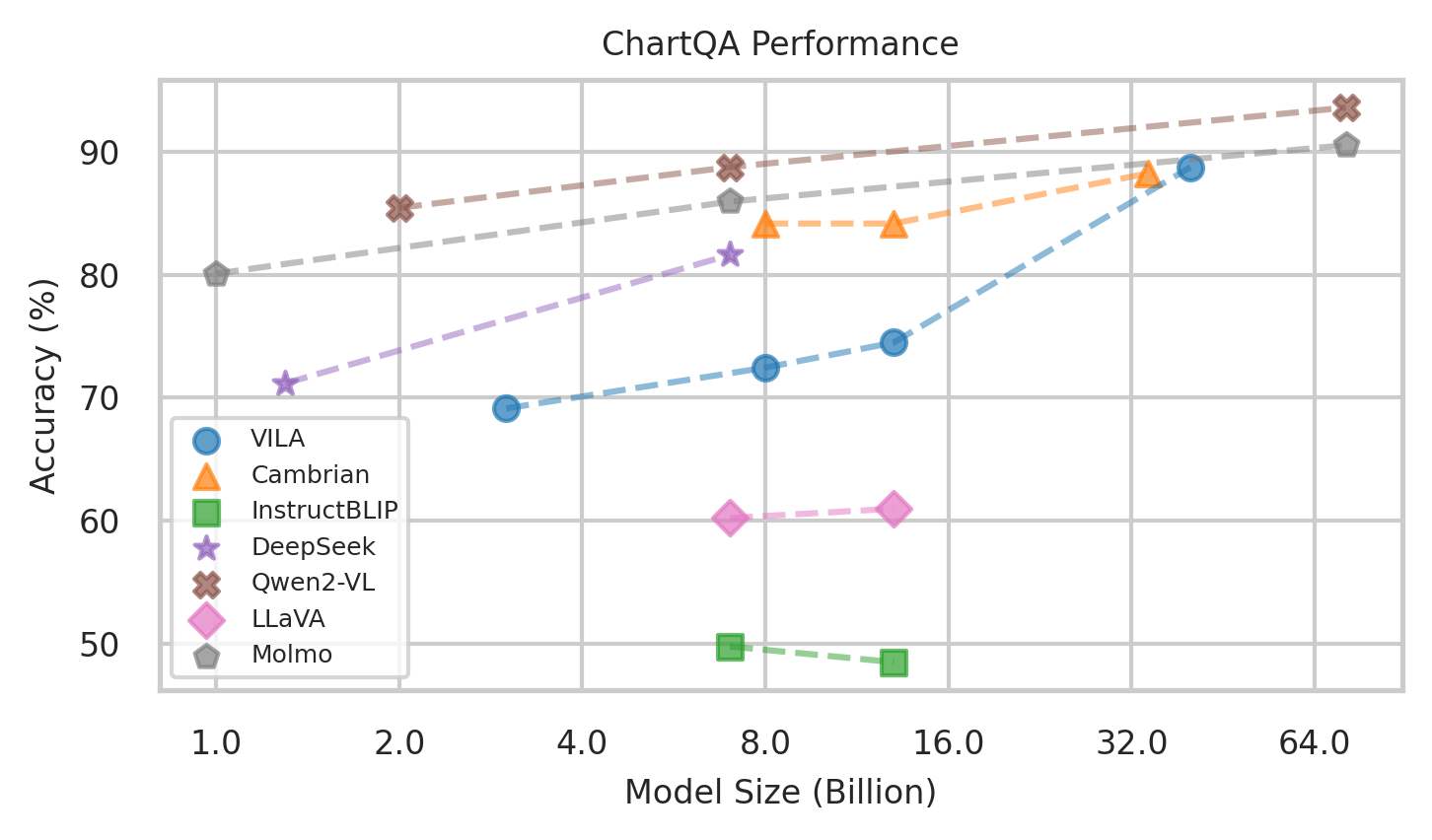}
\includegraphics[width=0.245\textwidth]{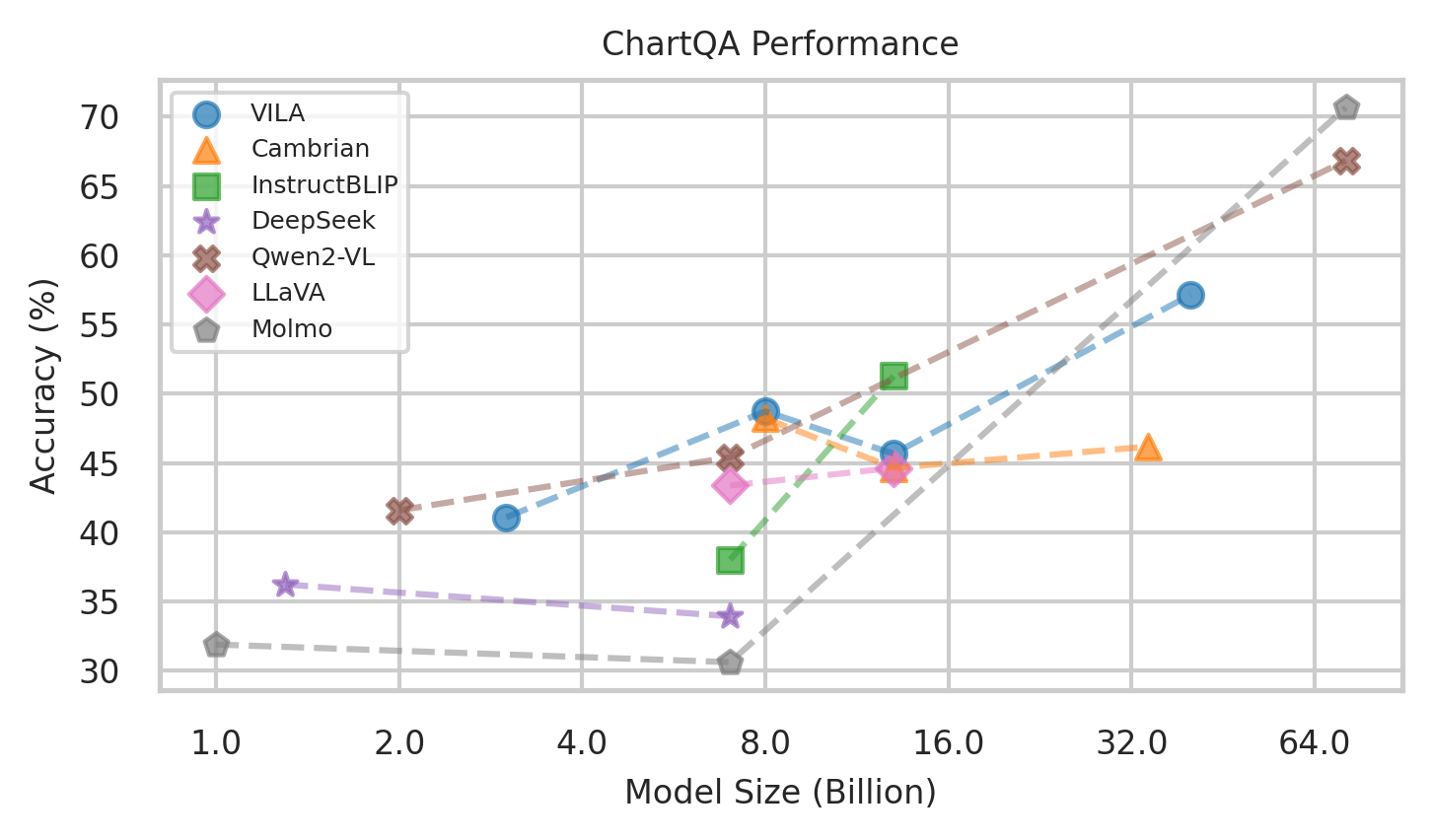}
\includegraphics[width=0.245\textwidth]{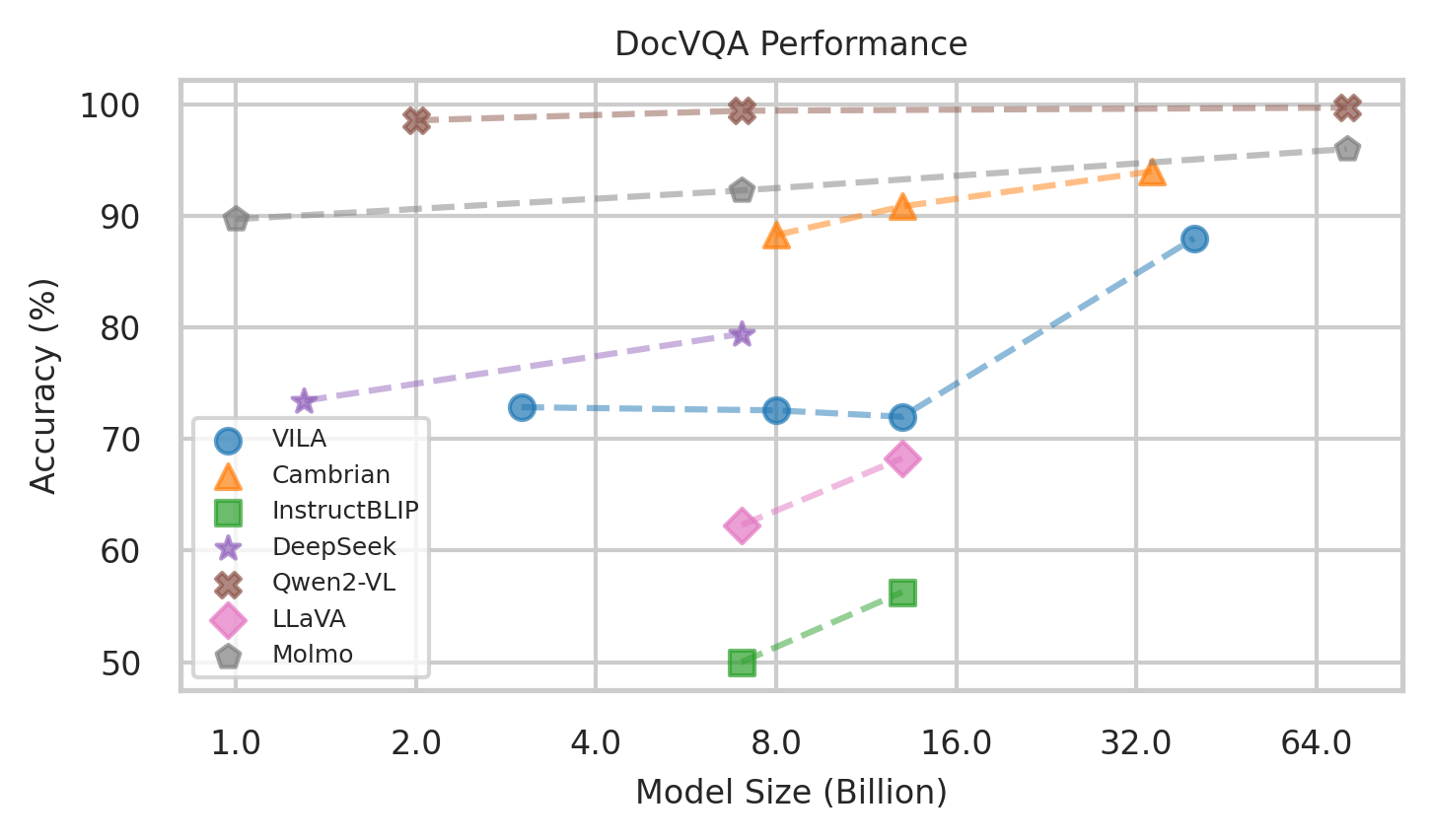}
\includegraphics[width=0.245\textwidth]{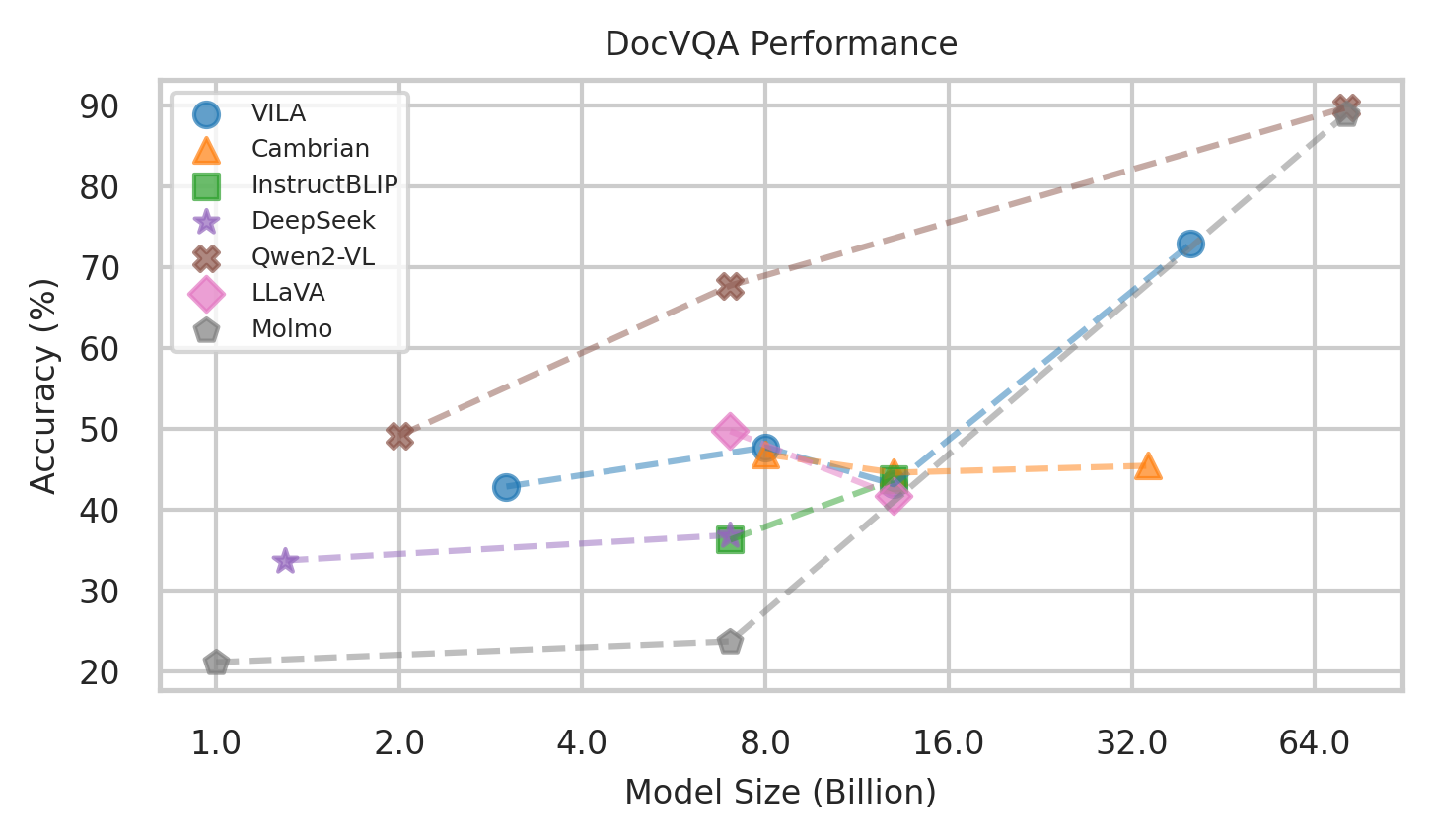}
\includegraphics[width=0.245\textwidth]{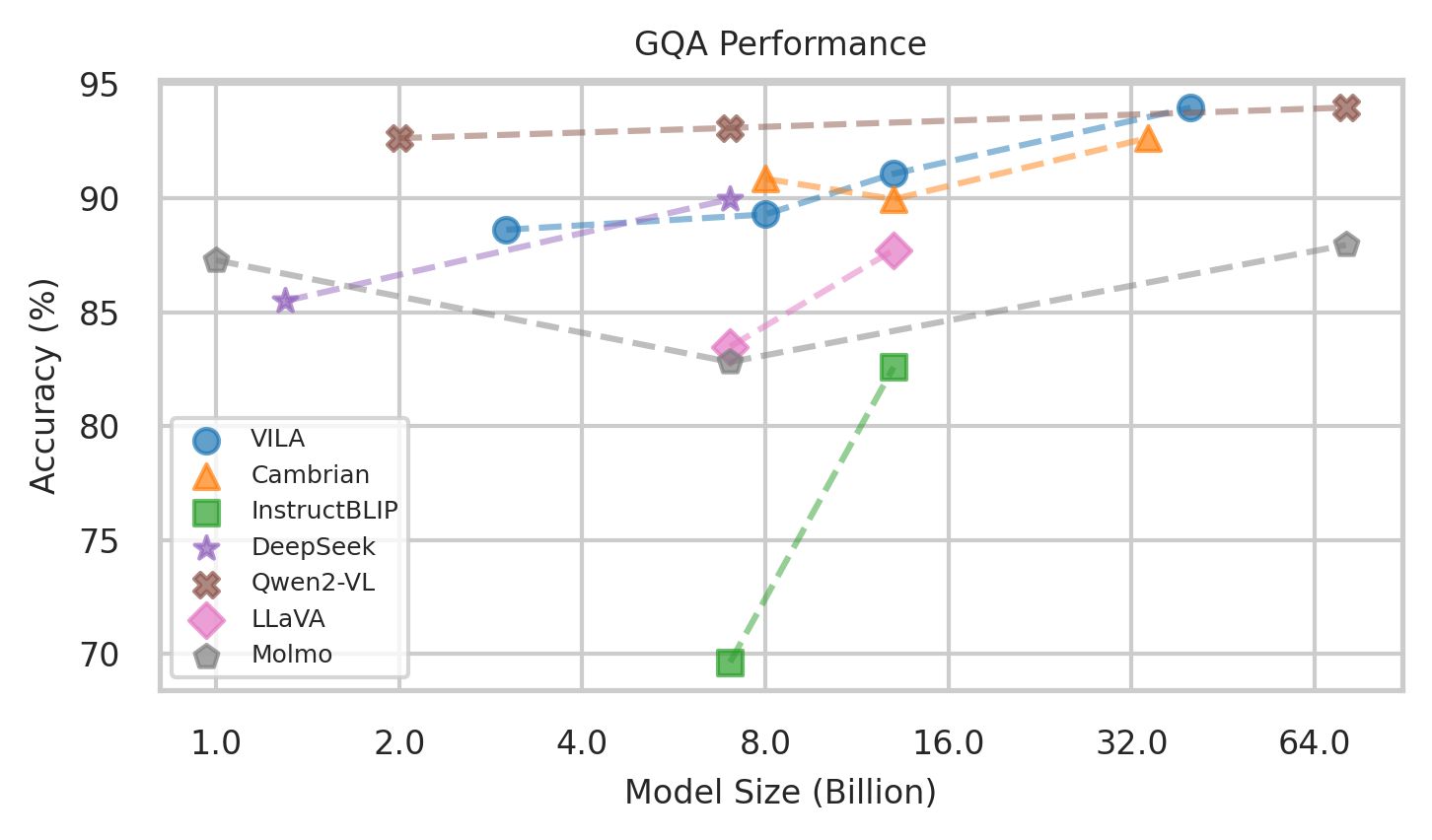}
\includegraphics[width=0.245\textwidth]{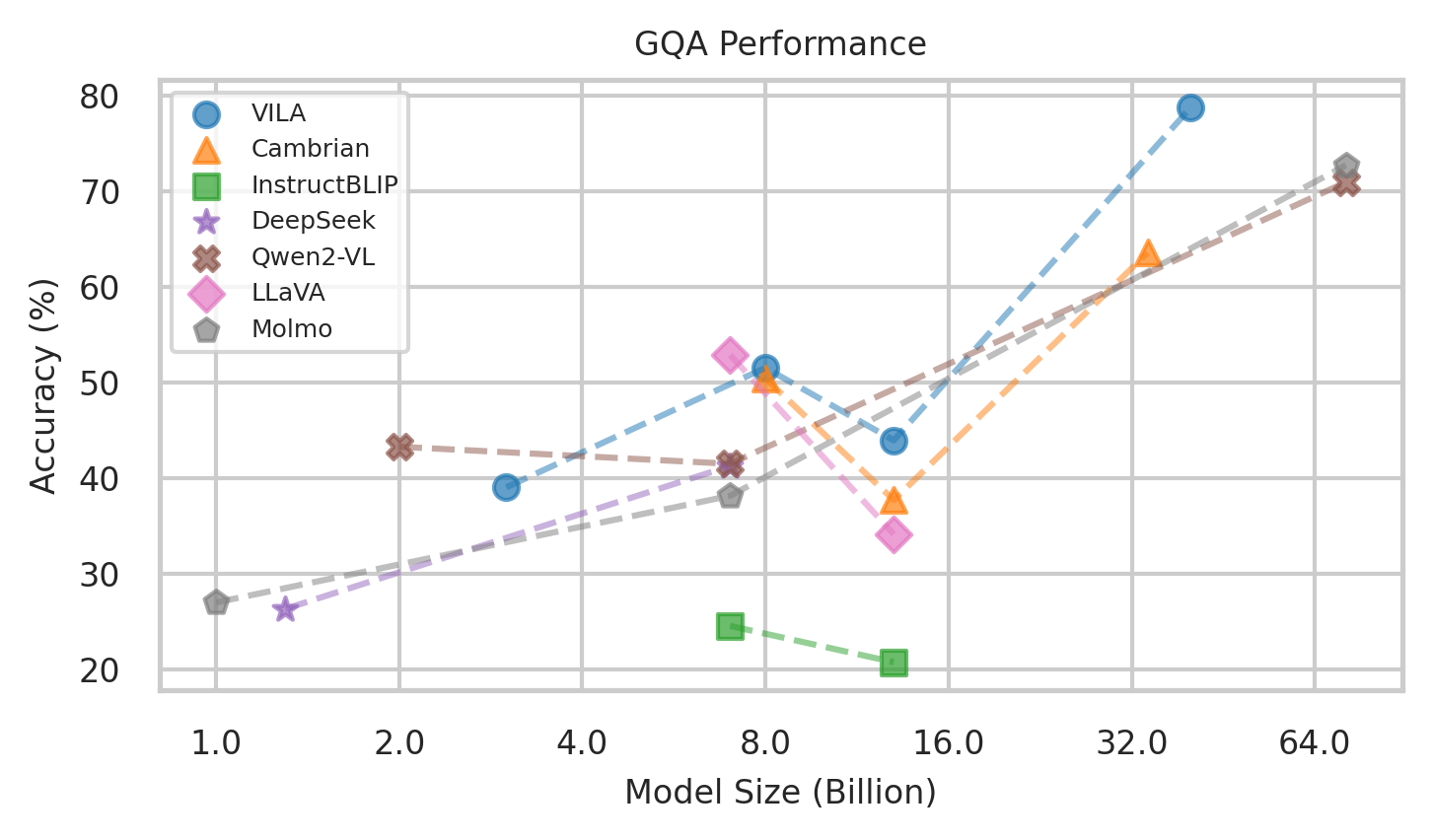}
\includegraphics[width=0.245\textwidth]{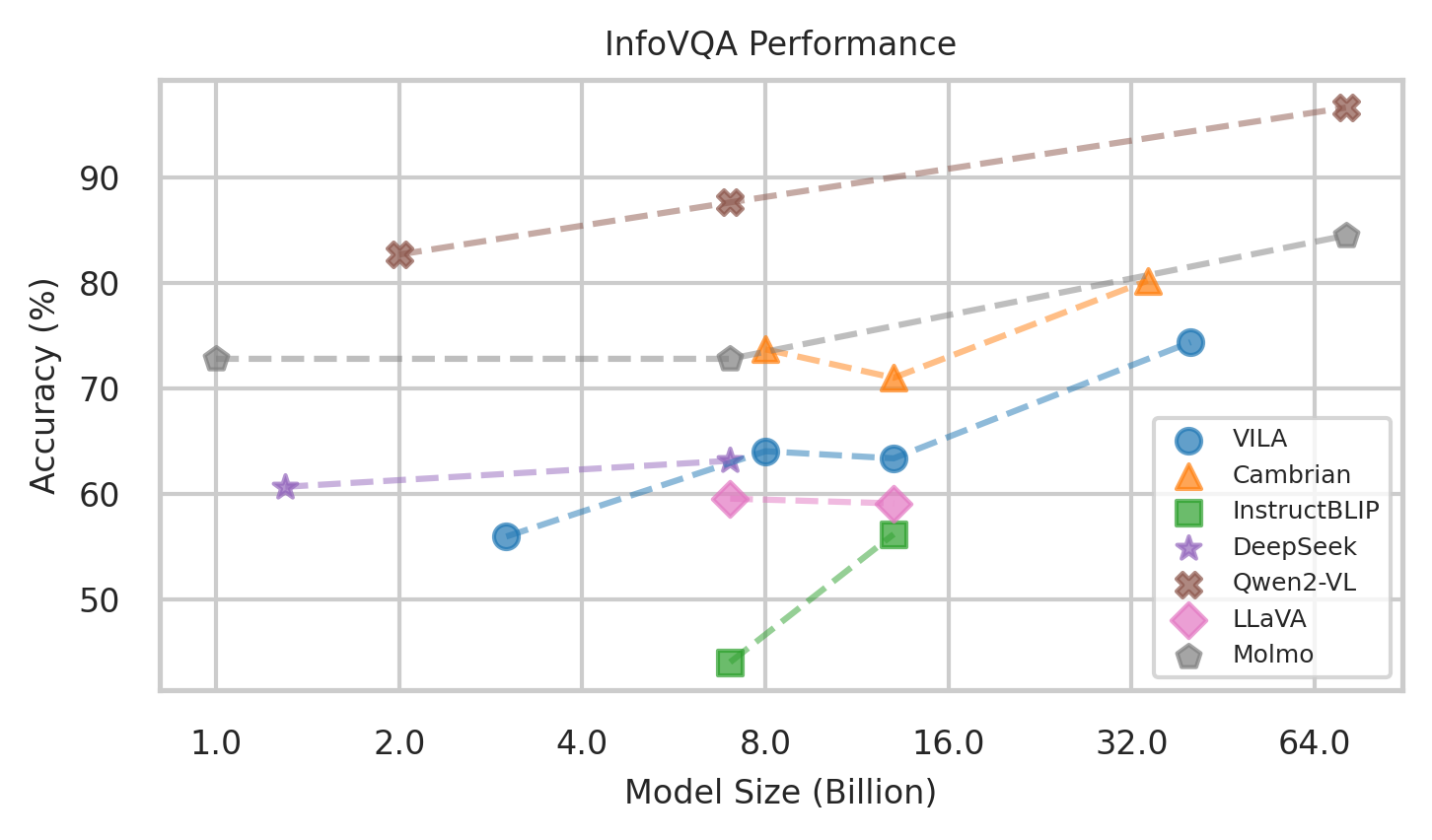}
\includegraphics[width=0.245\textwidth]{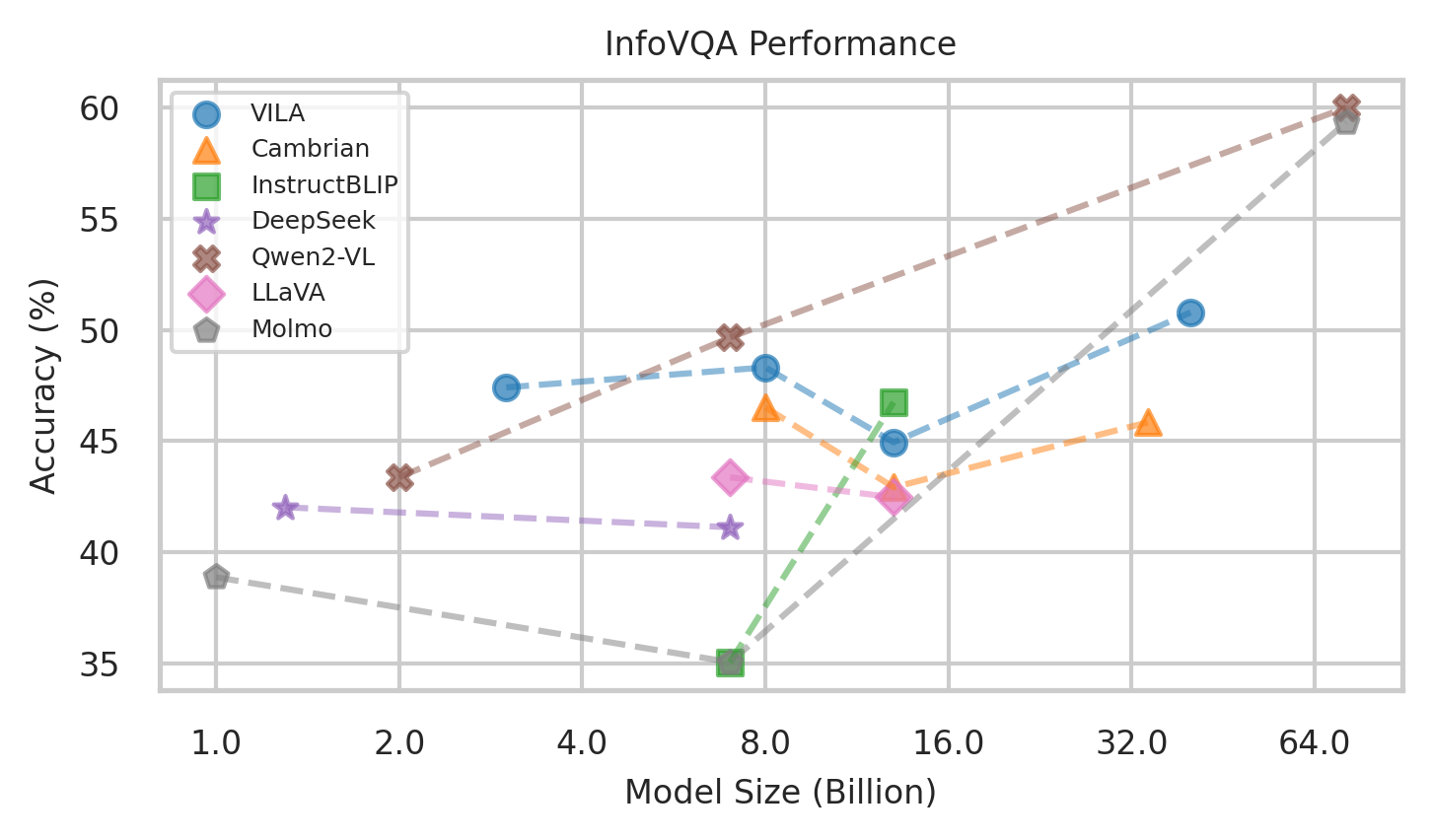}
\includegraphics[width=0.245\textwidth]{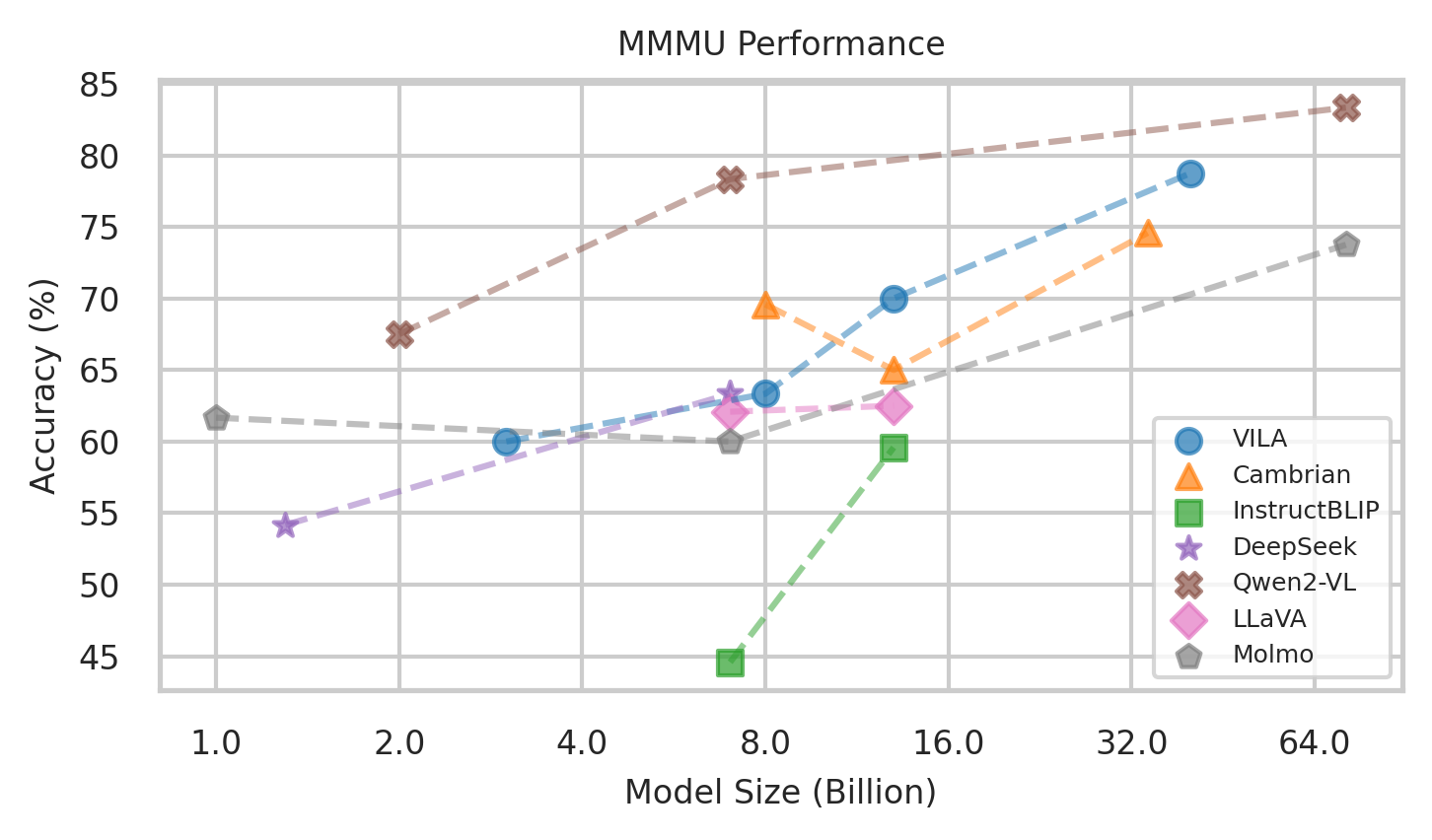}
\includegraphics[width=0.245\textwidth]{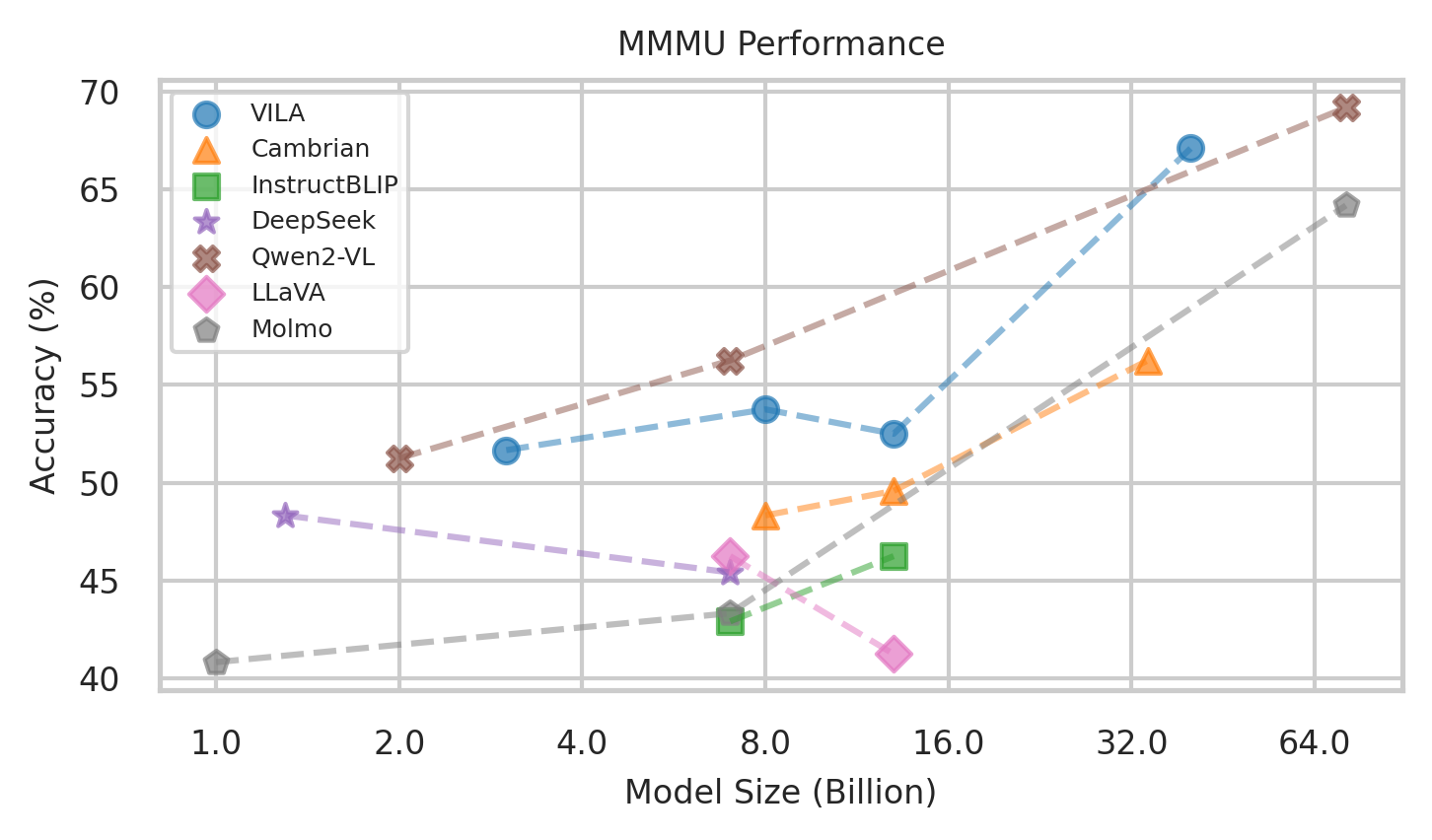}
\includegraphics[width=0.245\textwidth]{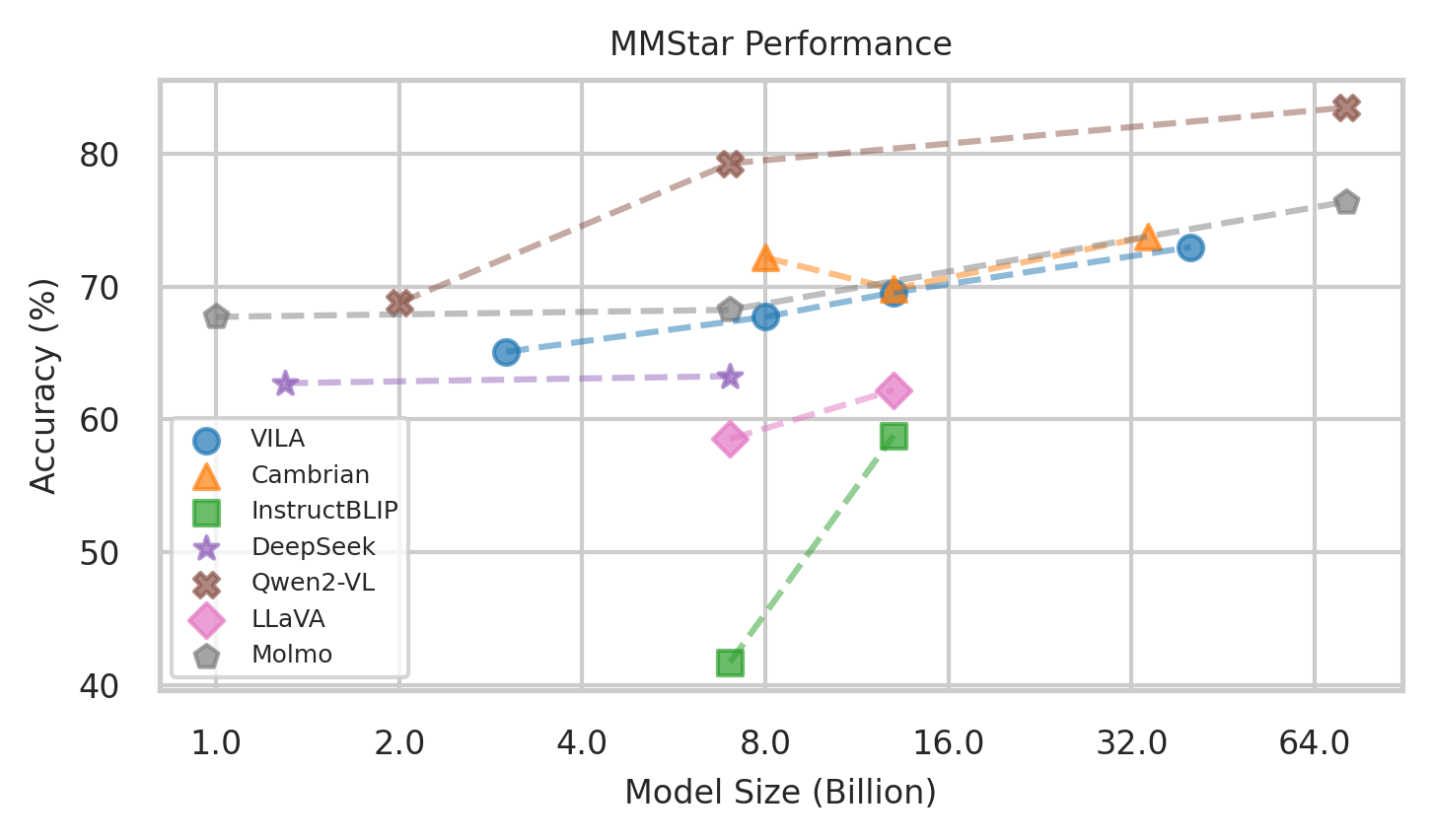}
\includegraphics[width=0.245\textwidth]{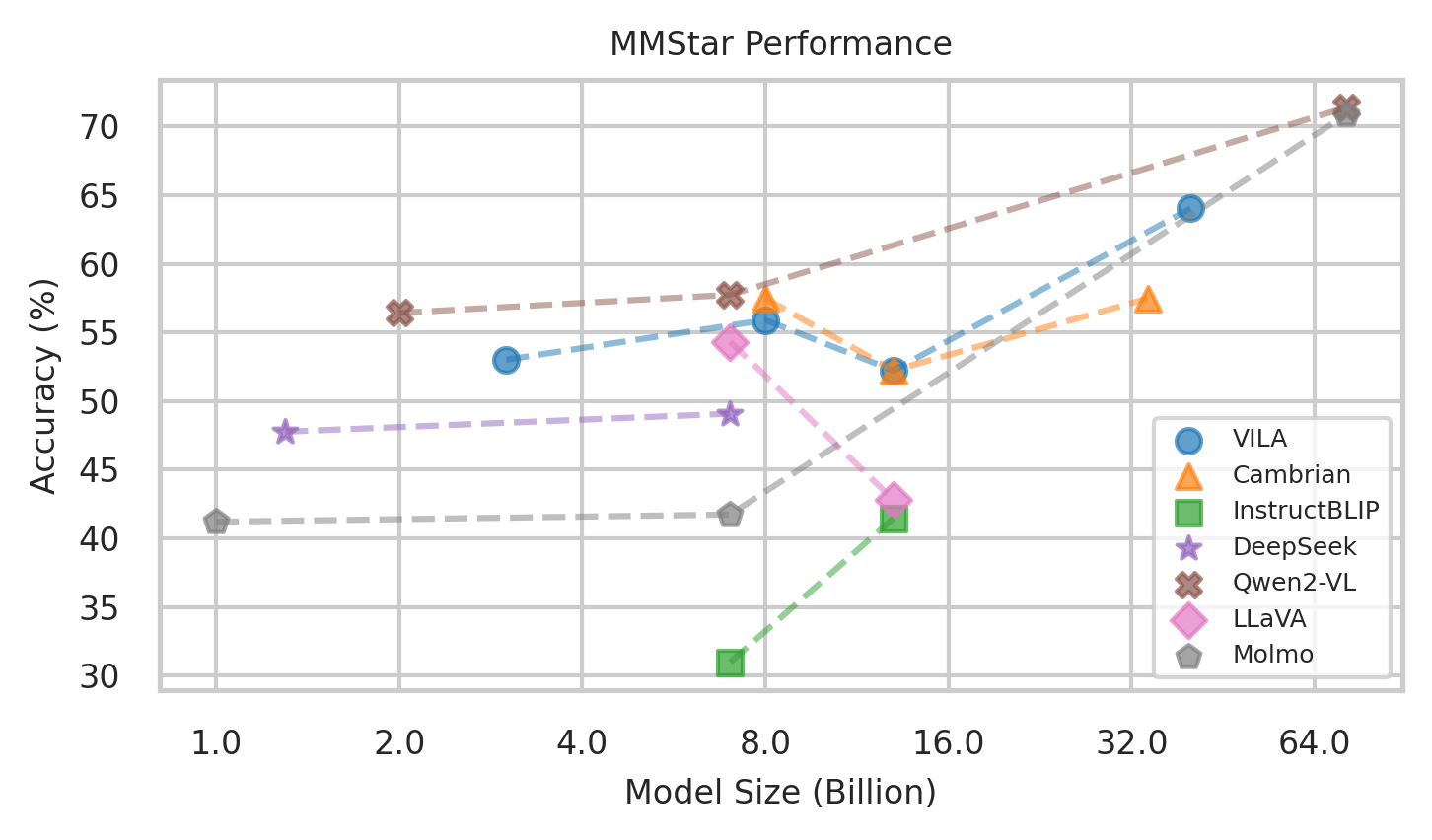}
\includegraphics[width=0.245\textwidth]{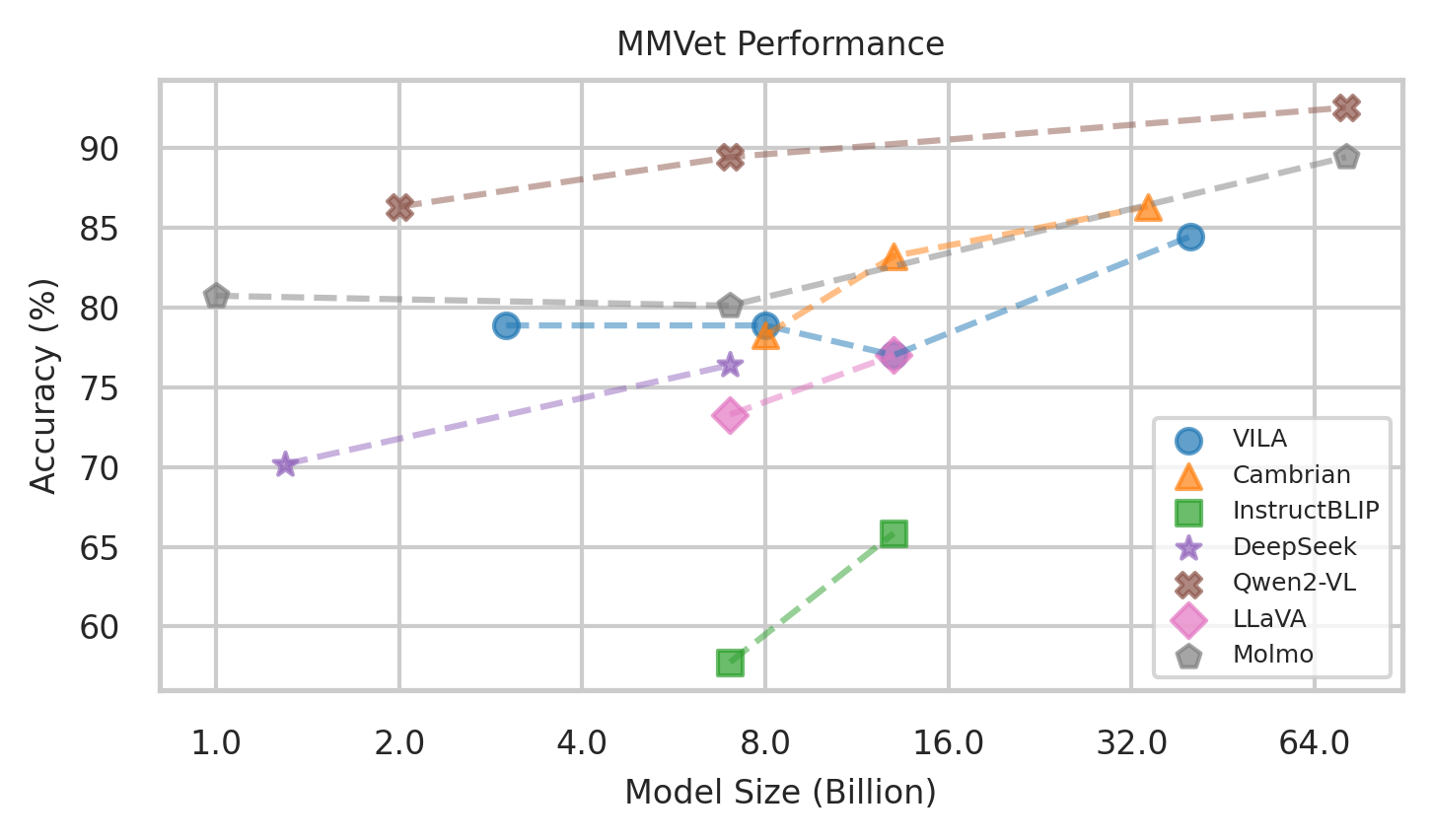}
\includegraphics[width=0.245\textwidth]{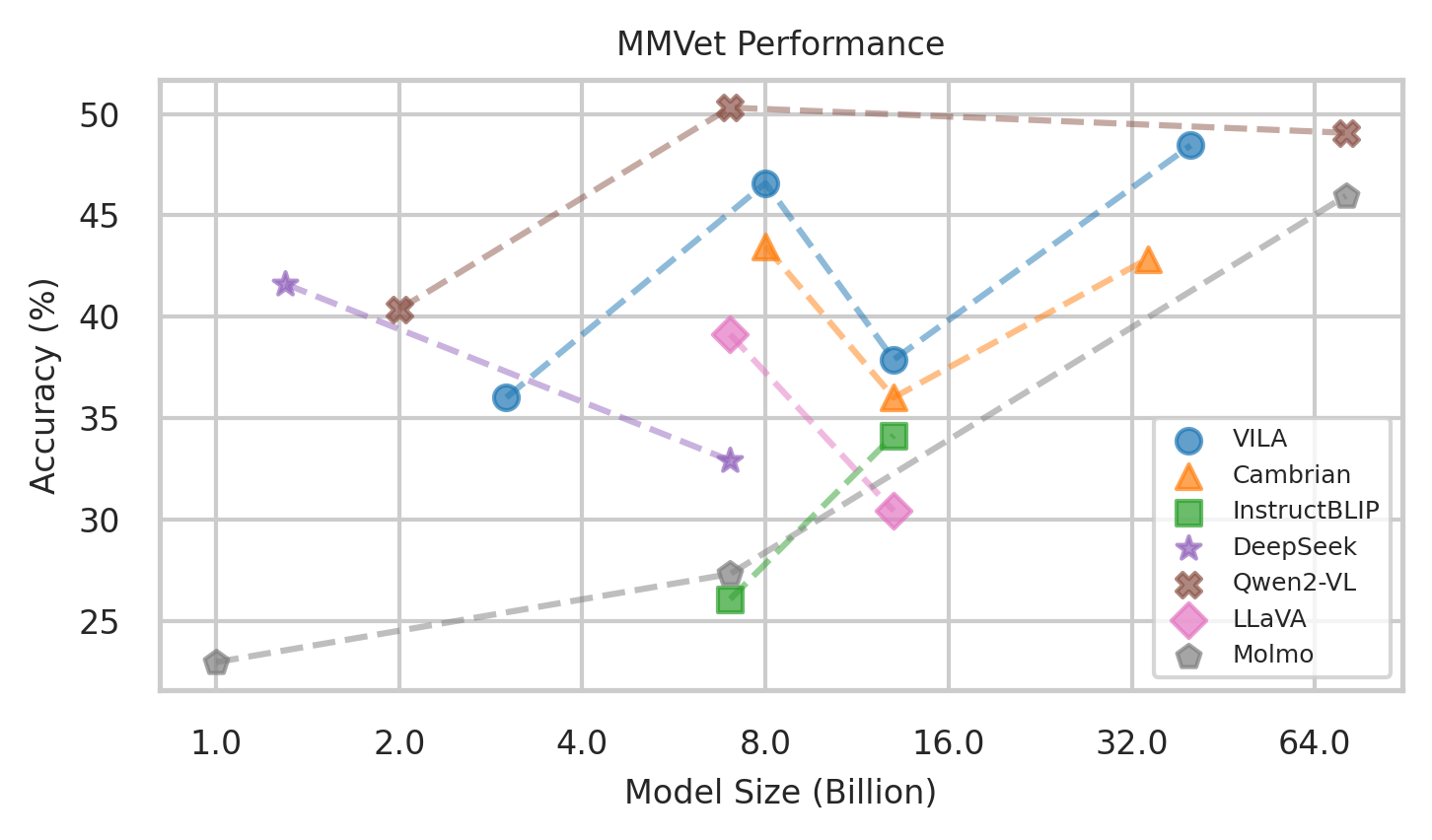}
\includegraphics[width=0.245\textwidth]{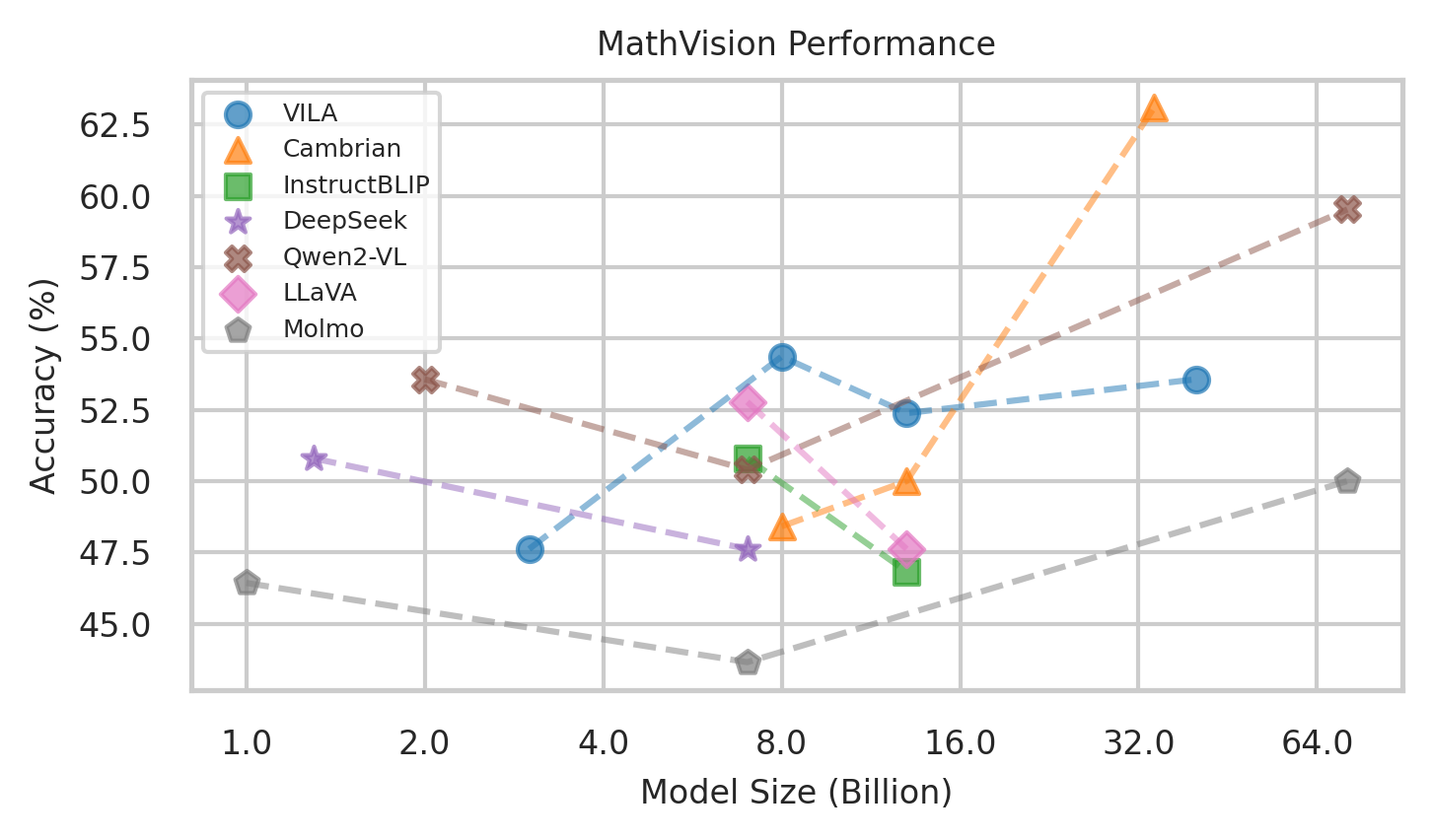}
\includegraphics[width=0.245\textwidth]{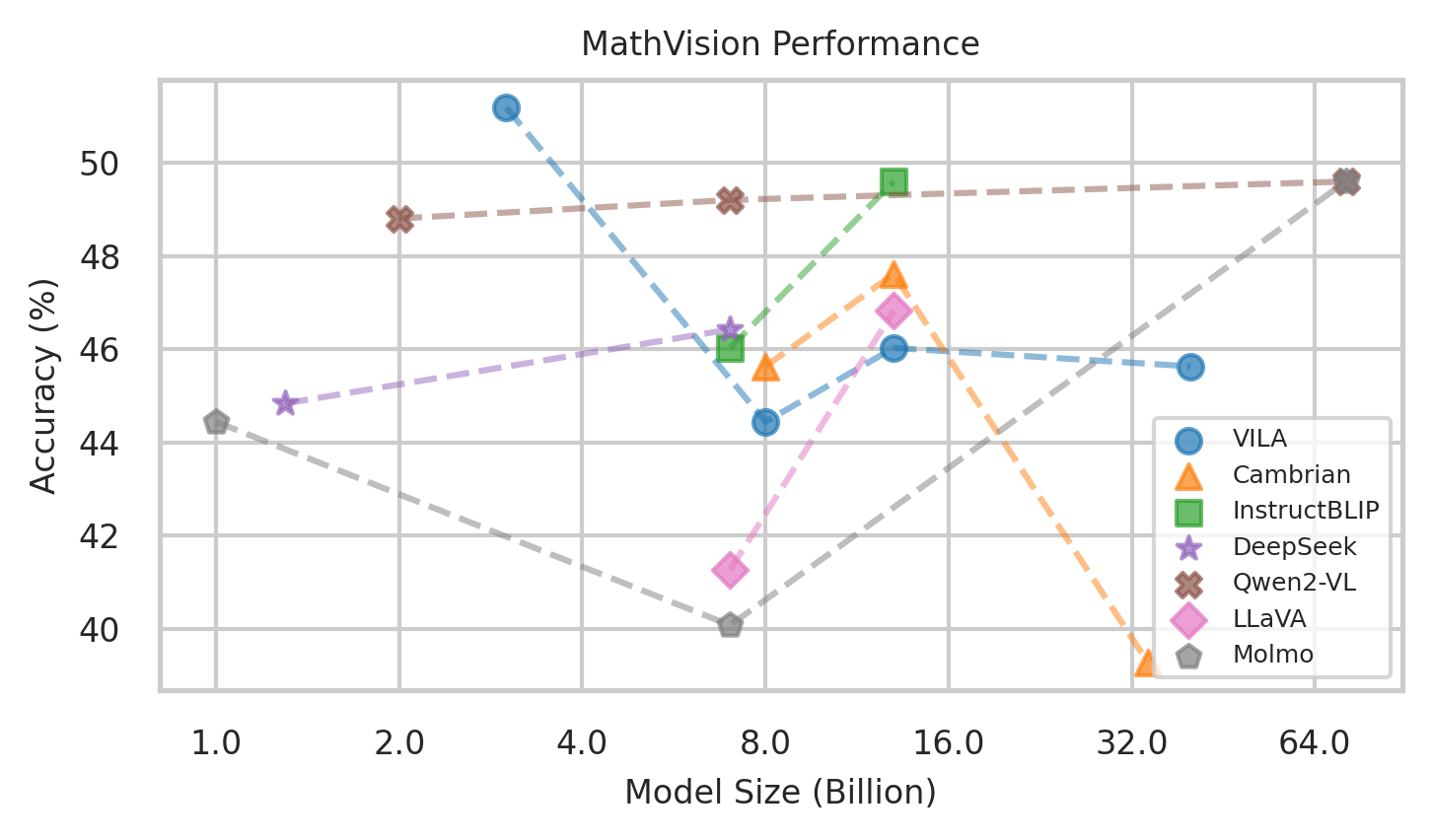}
\includegraphics[width=0.245\textwidth]{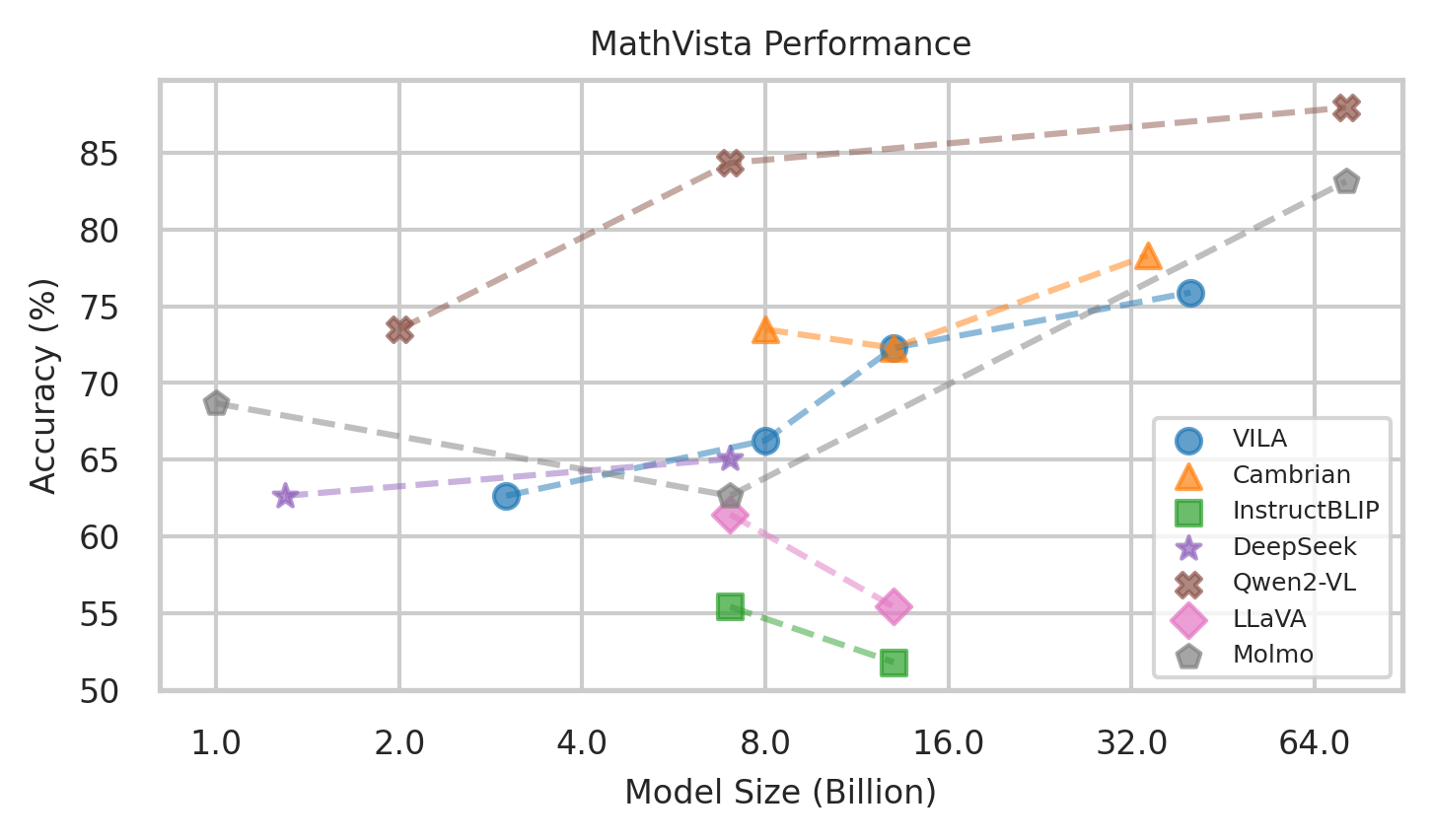}
\includegraphics[width=0.245\textwidth]{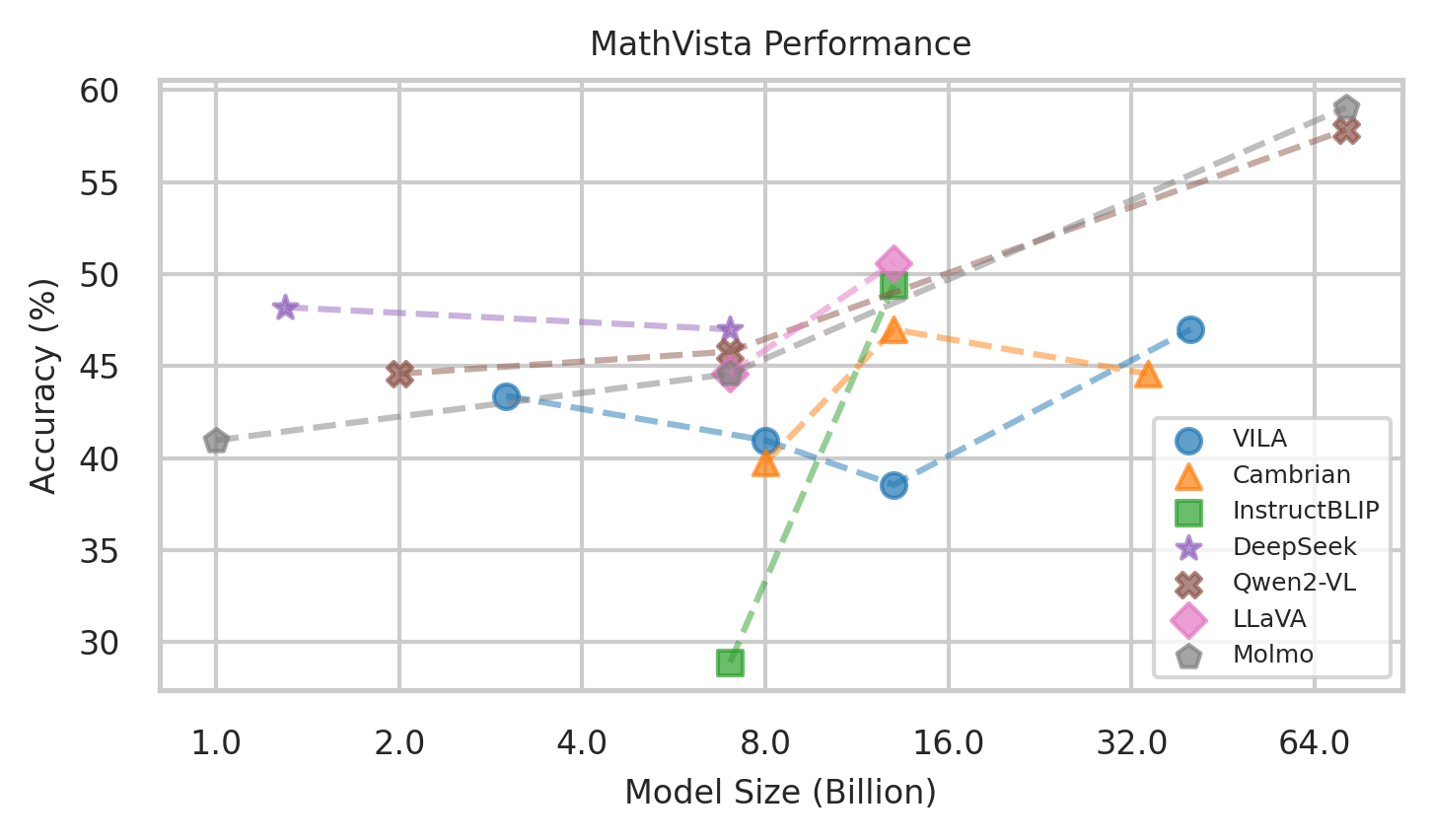}
\includegraphics[width=0.245\textwidth]{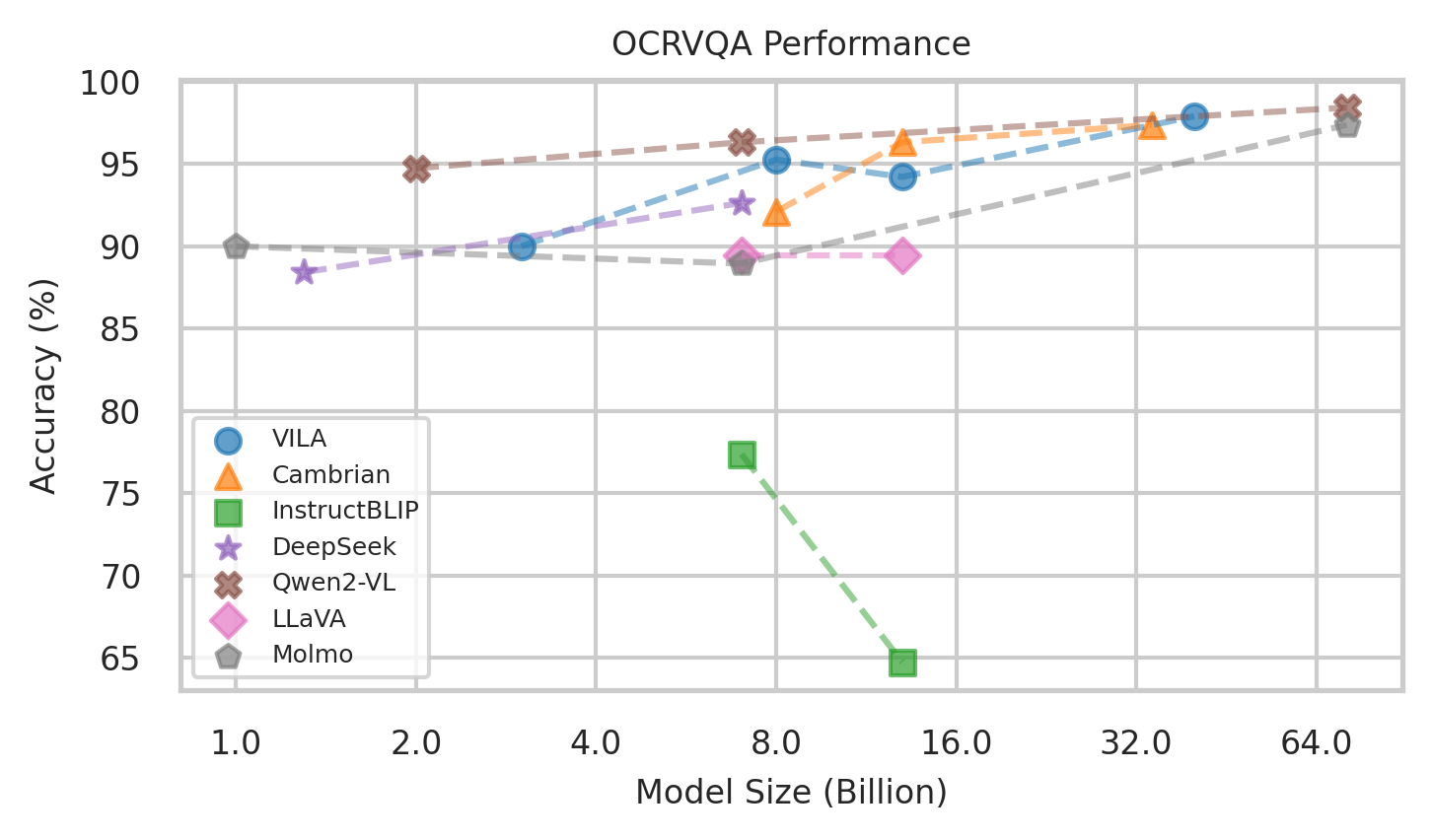}
\includegraphics[width=0.245\textwidth]{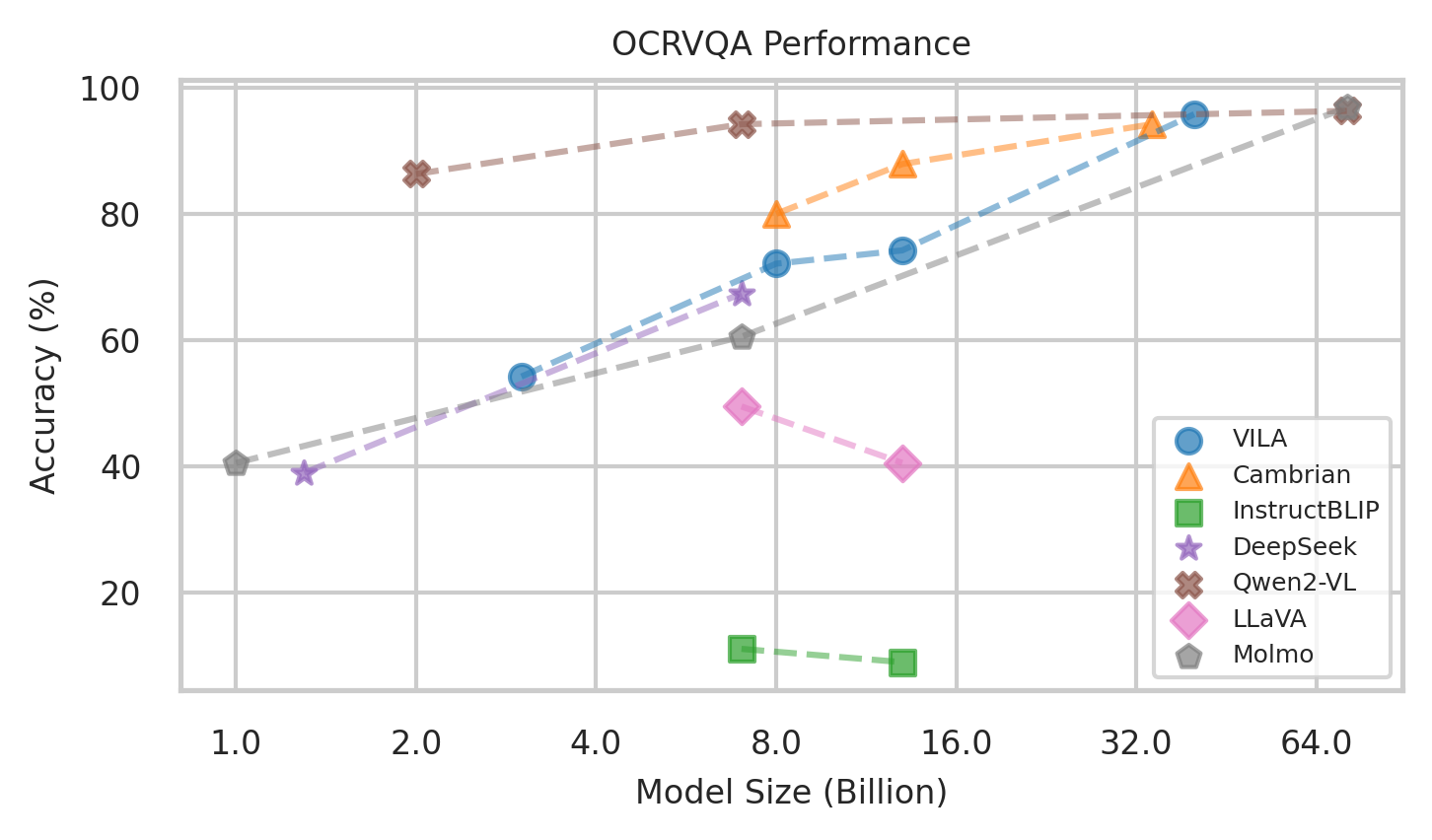}
\includegraphics[width=0.245\textwidth]{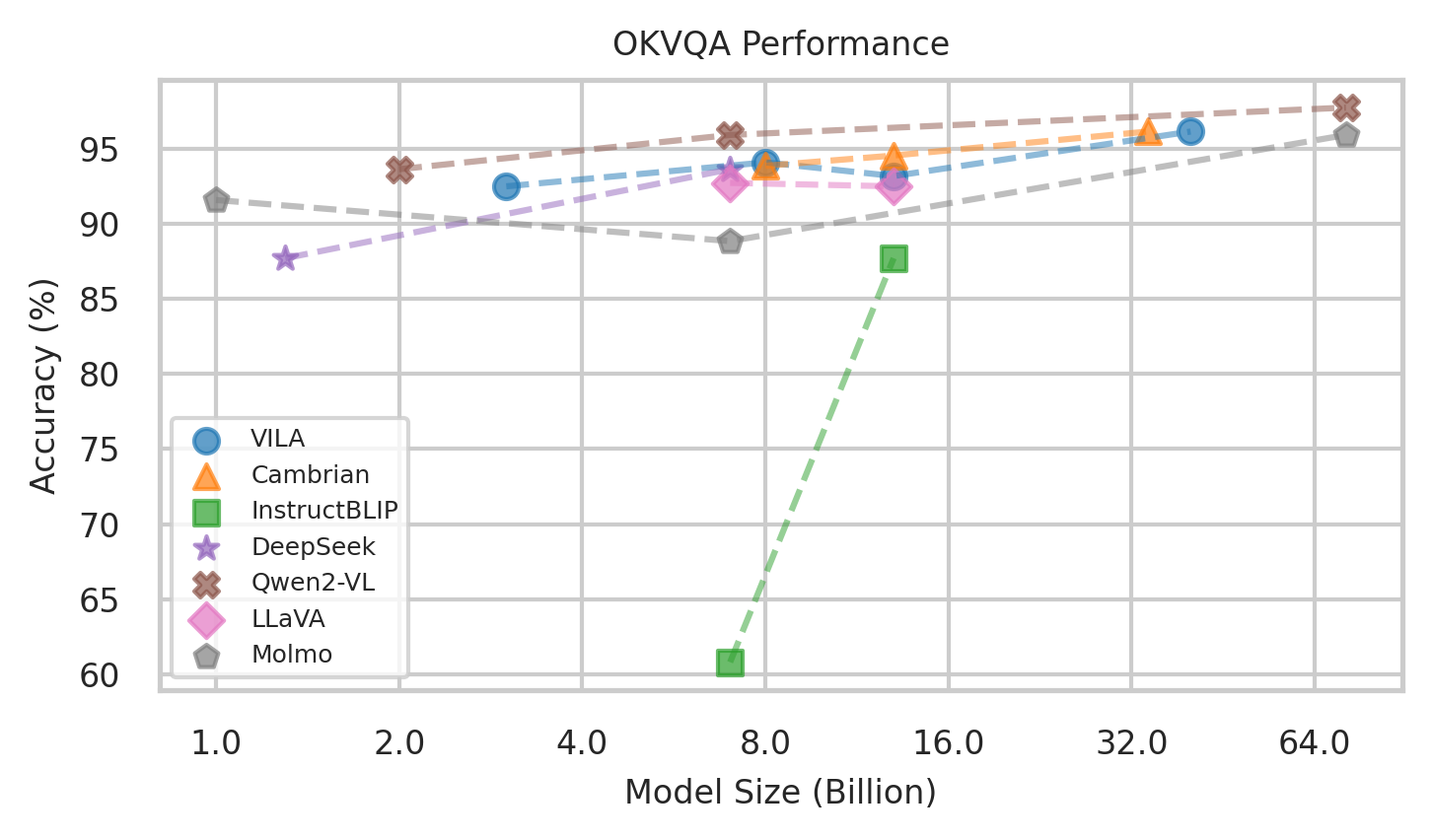}
\includegraphics[width=0.245\textwidth]{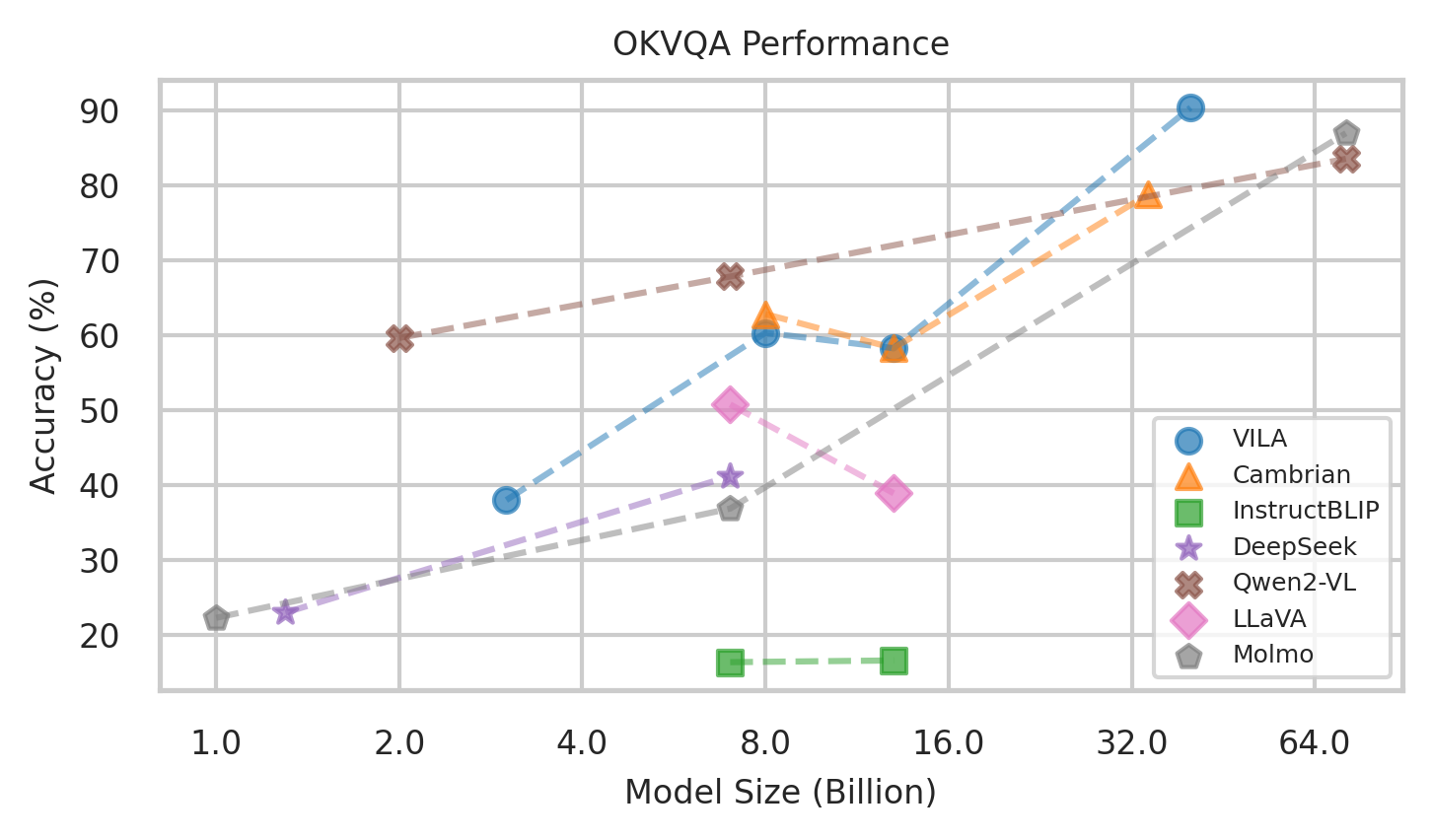}
\includegraphics[width=0.245\textwidth]{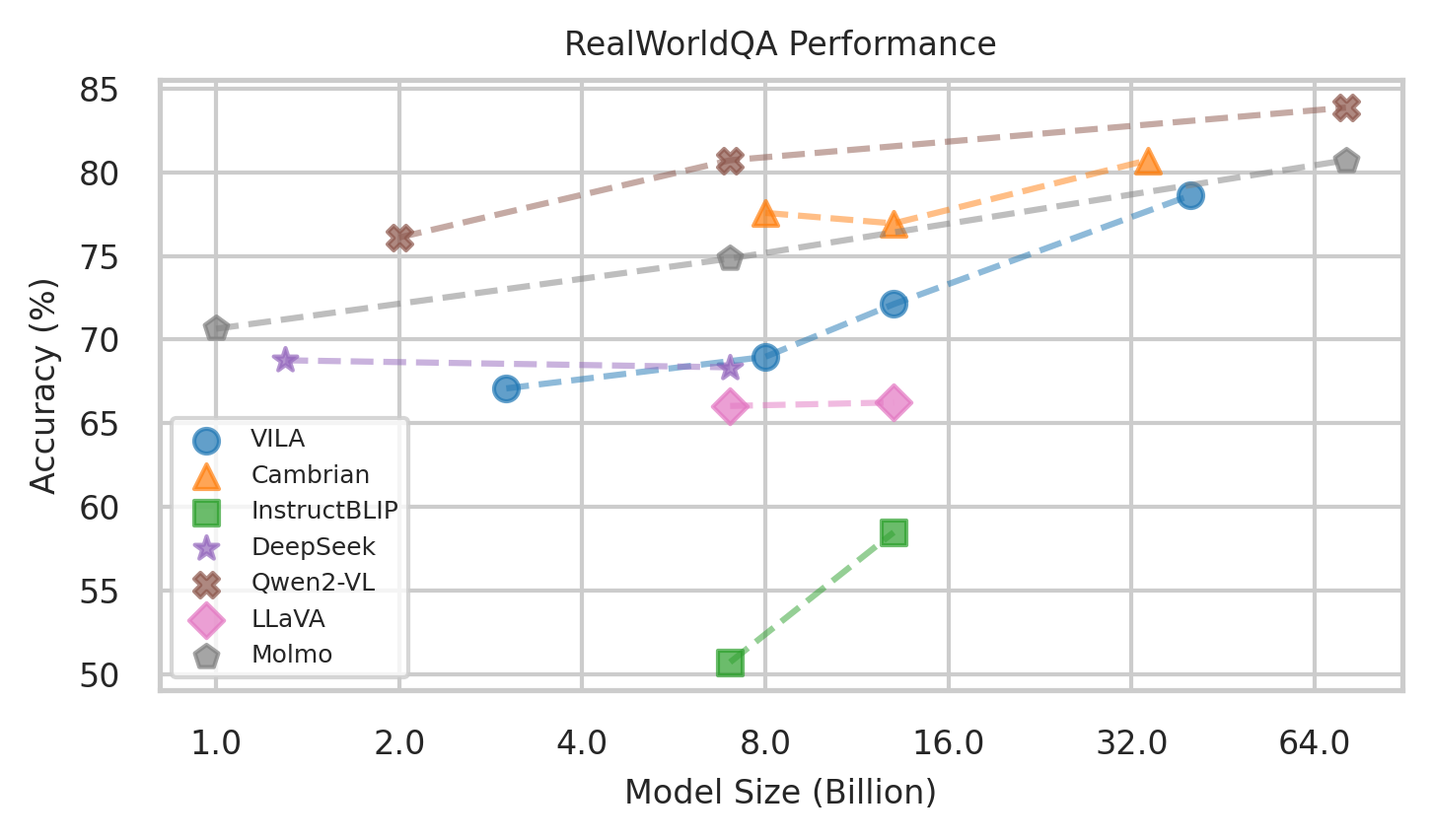}
\includegraphics[width=0.245\textwidth]{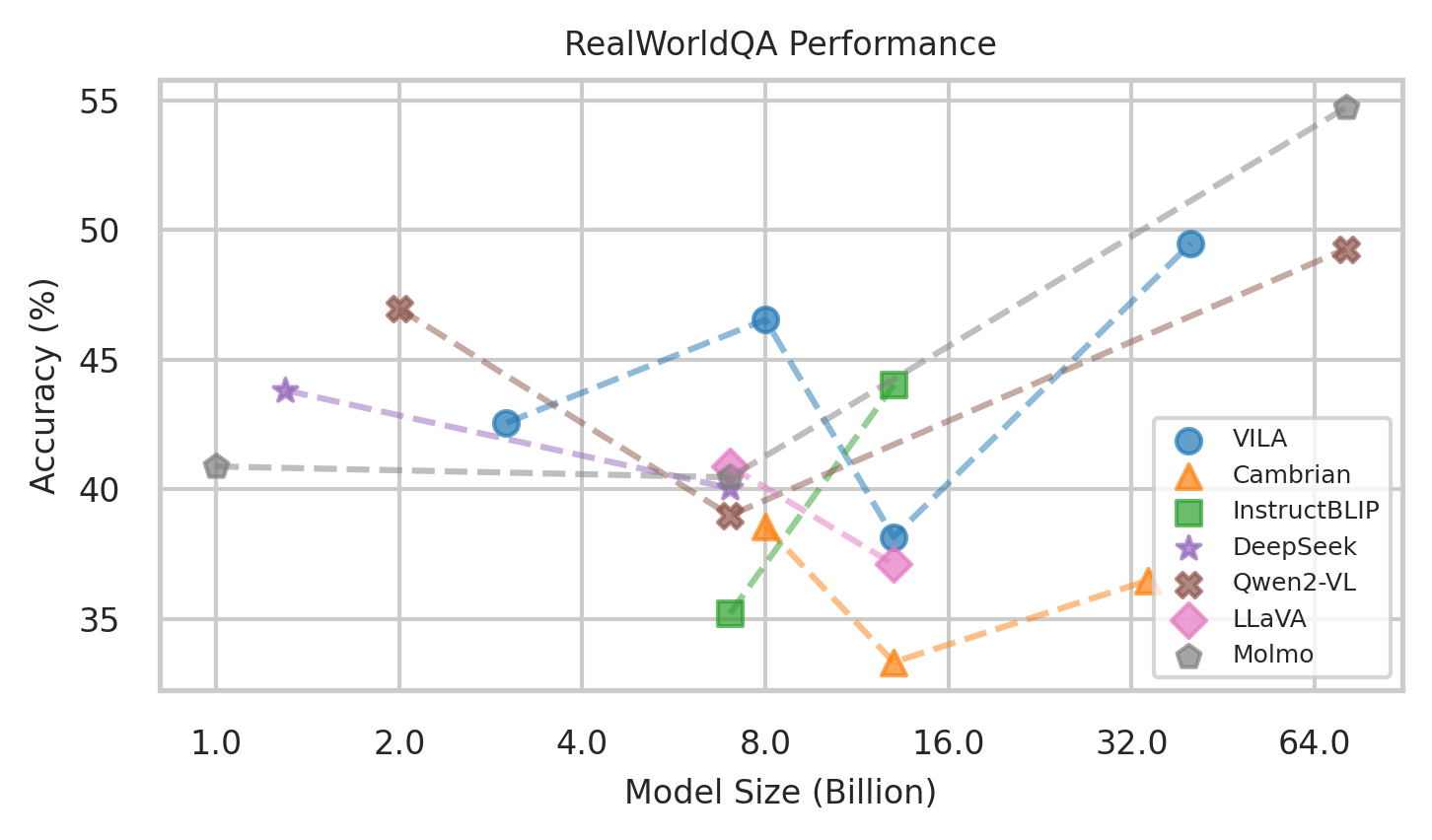}
\includegraphics[width=0.245\textwidth]{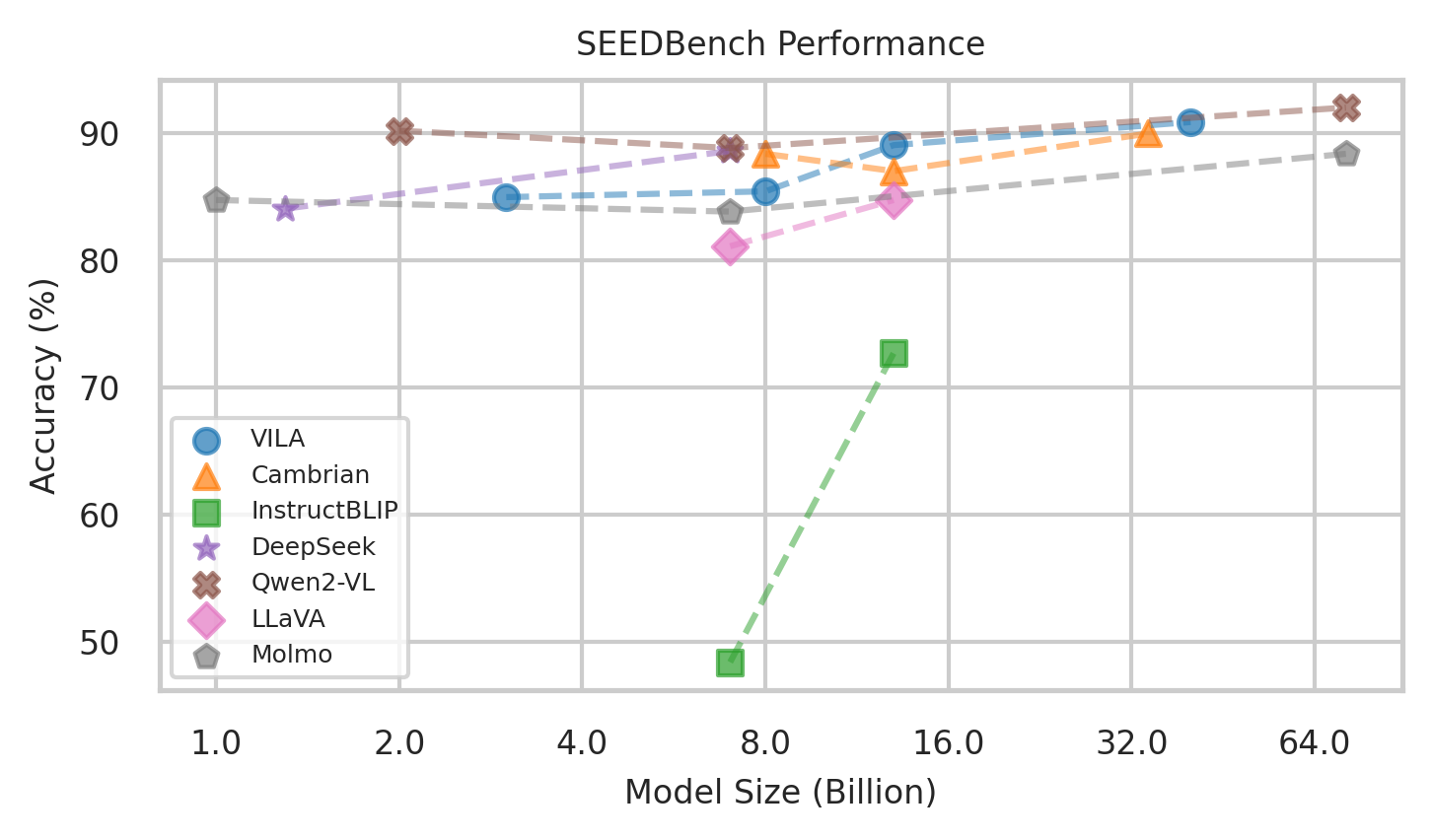}
\includegraphics[width=0.245\textwidth]{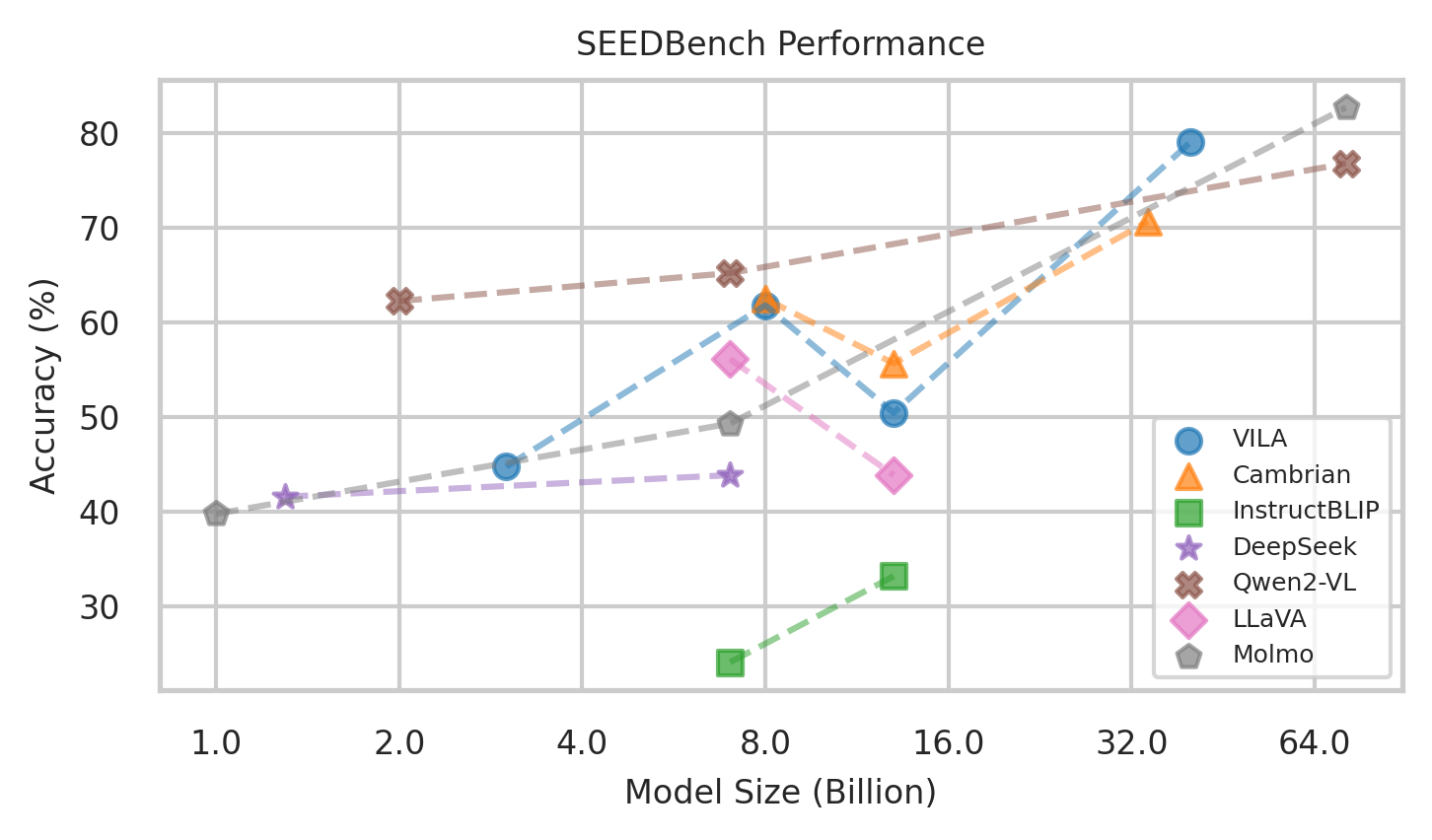}
\includegraphics[width=0.245\textwidth]{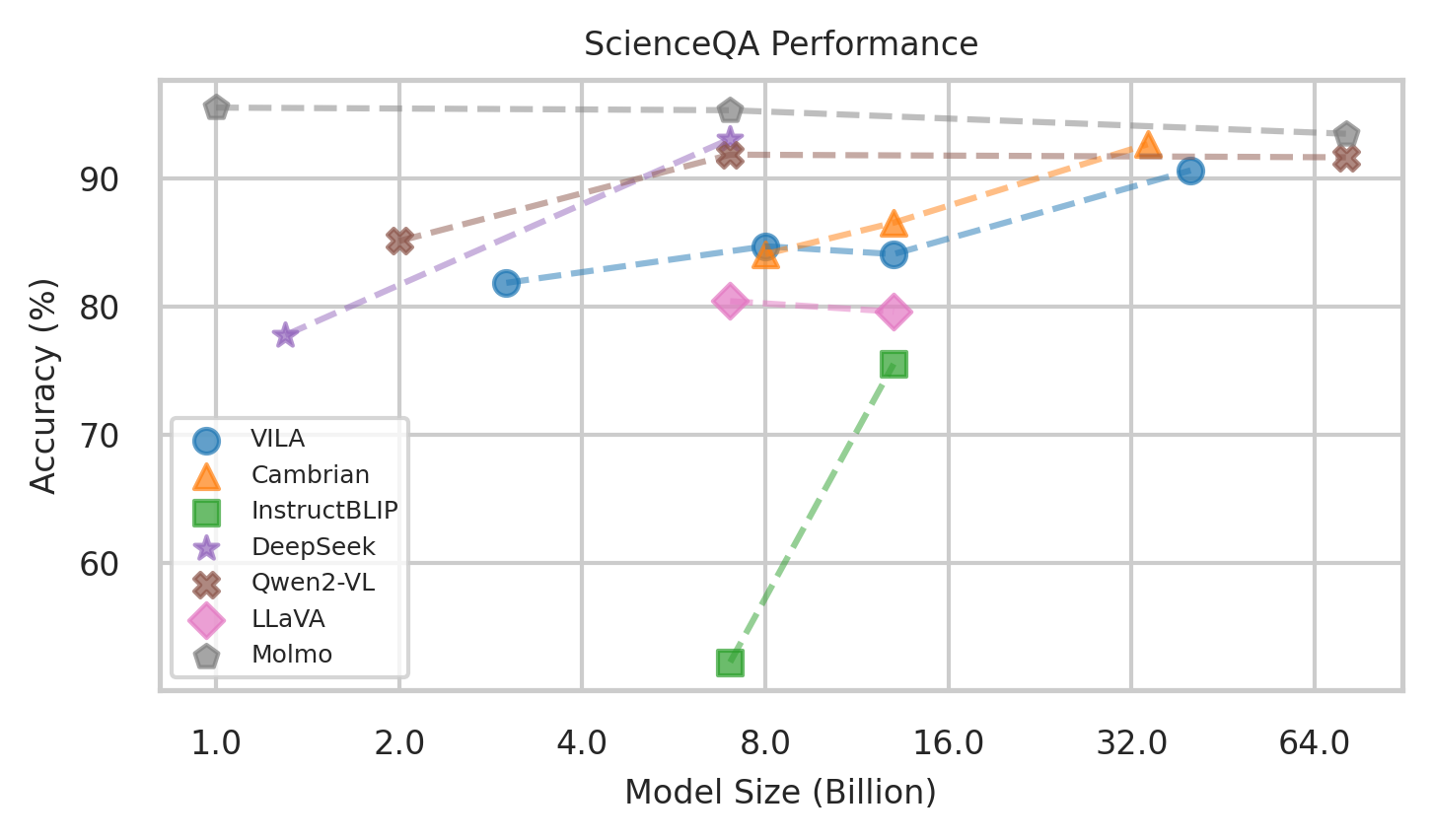}
\includegraphics[width=0.245\textwidth]{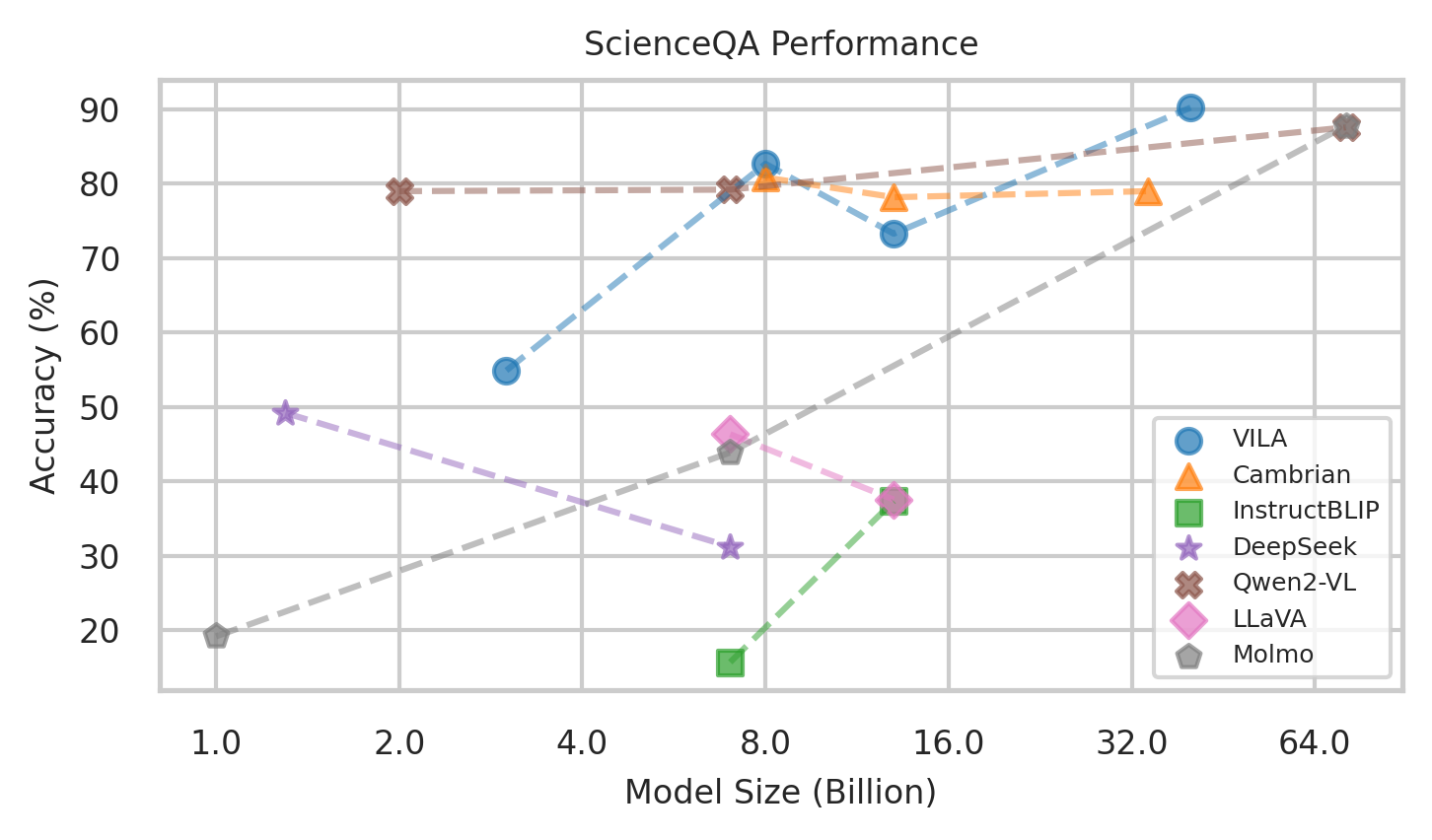}
\includegraphics[width=0.245\textwidth]{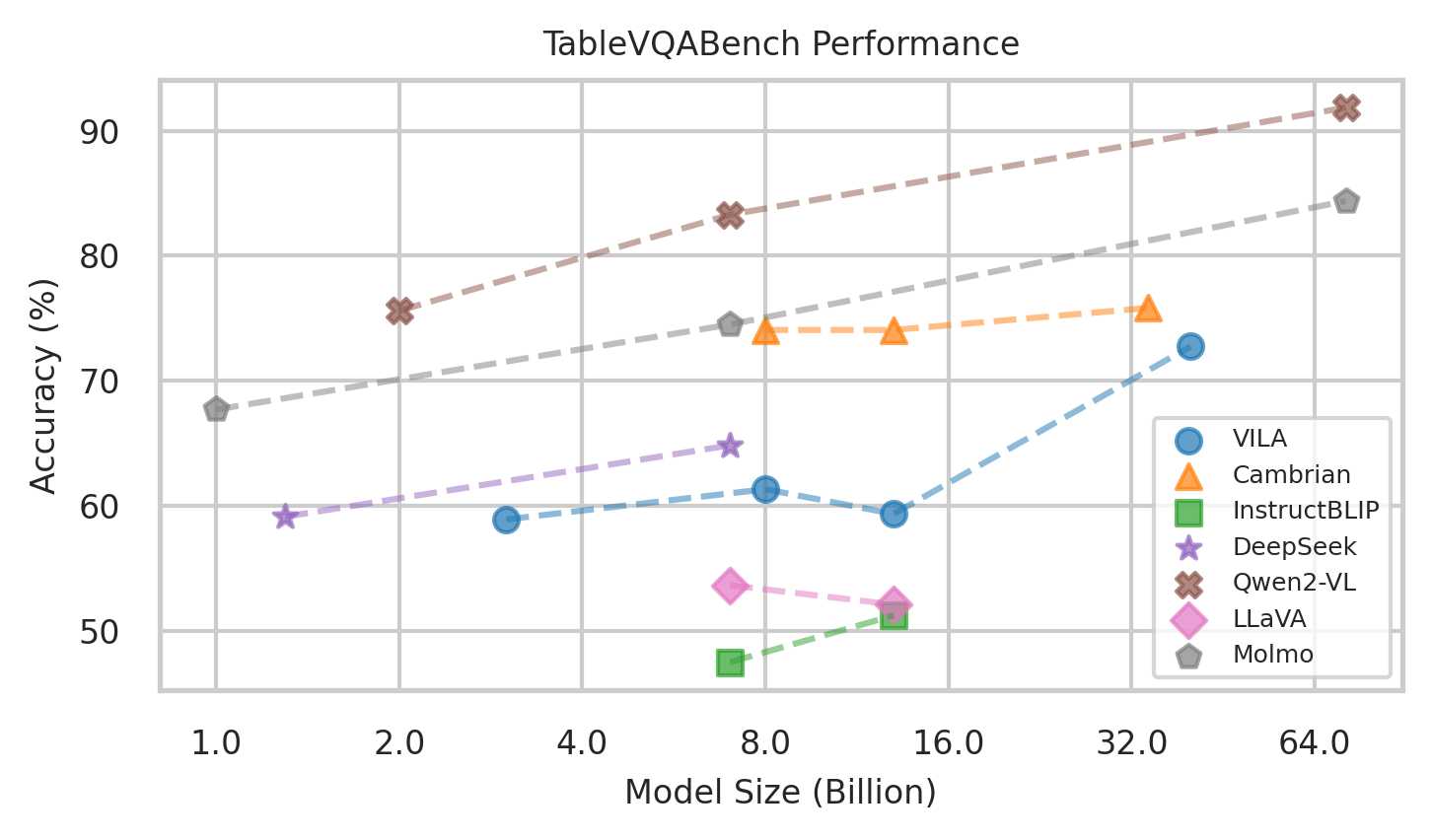}
\includegraphics[width=0.245\textwidth]{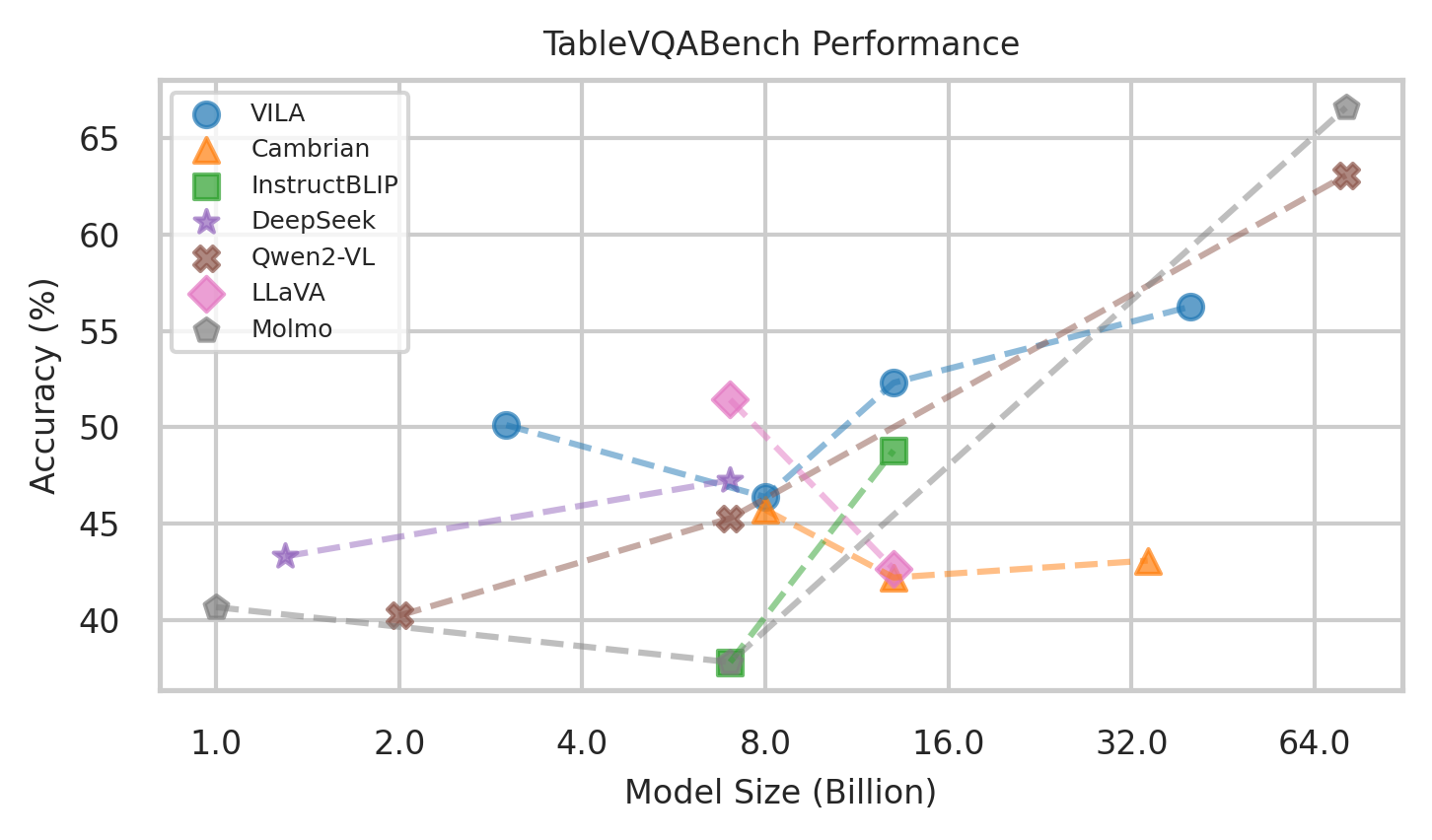}
\includegraphics[width=0.245\textwidth]{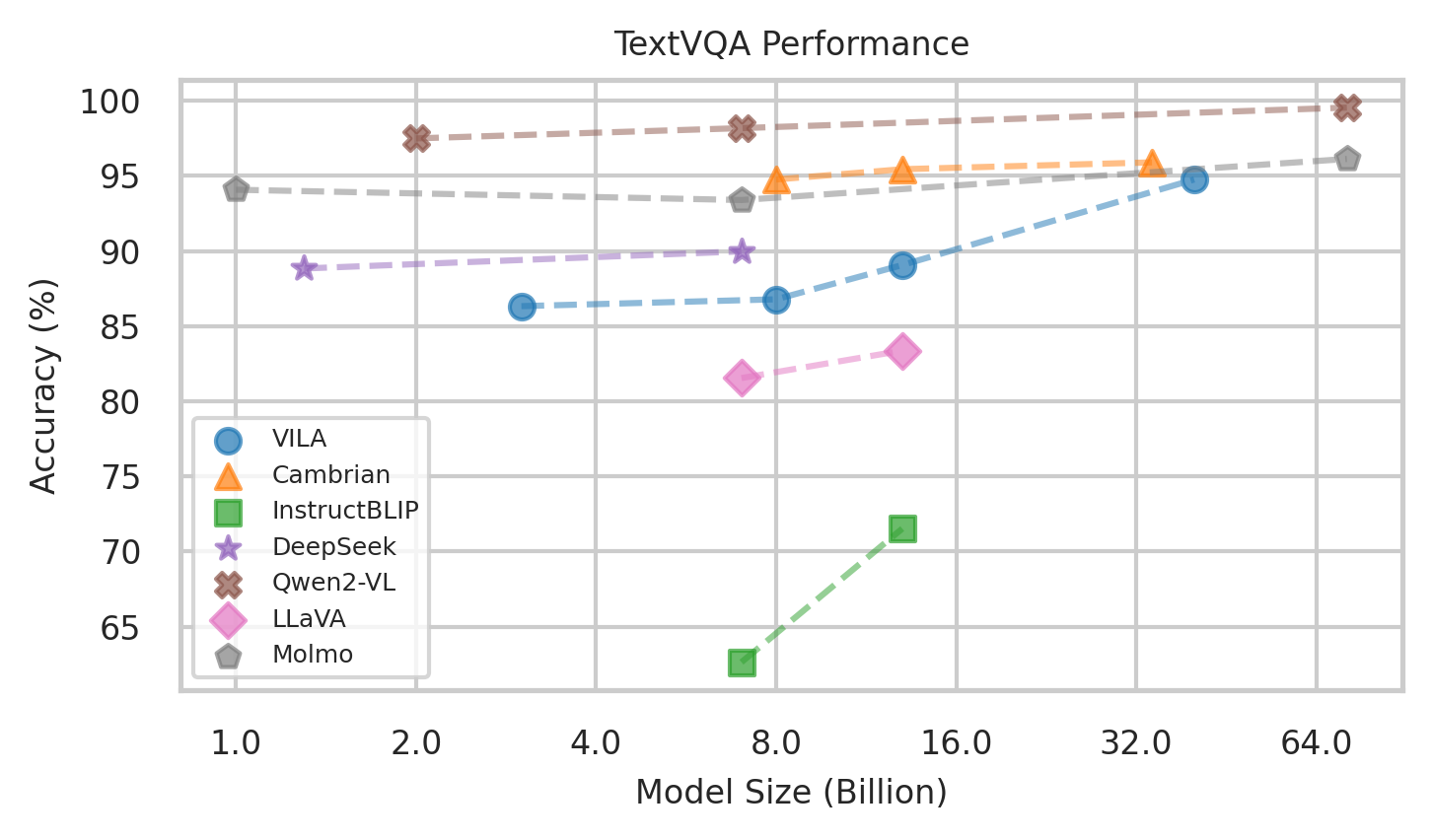}
\includegraphics[width=0.245\textwidth]{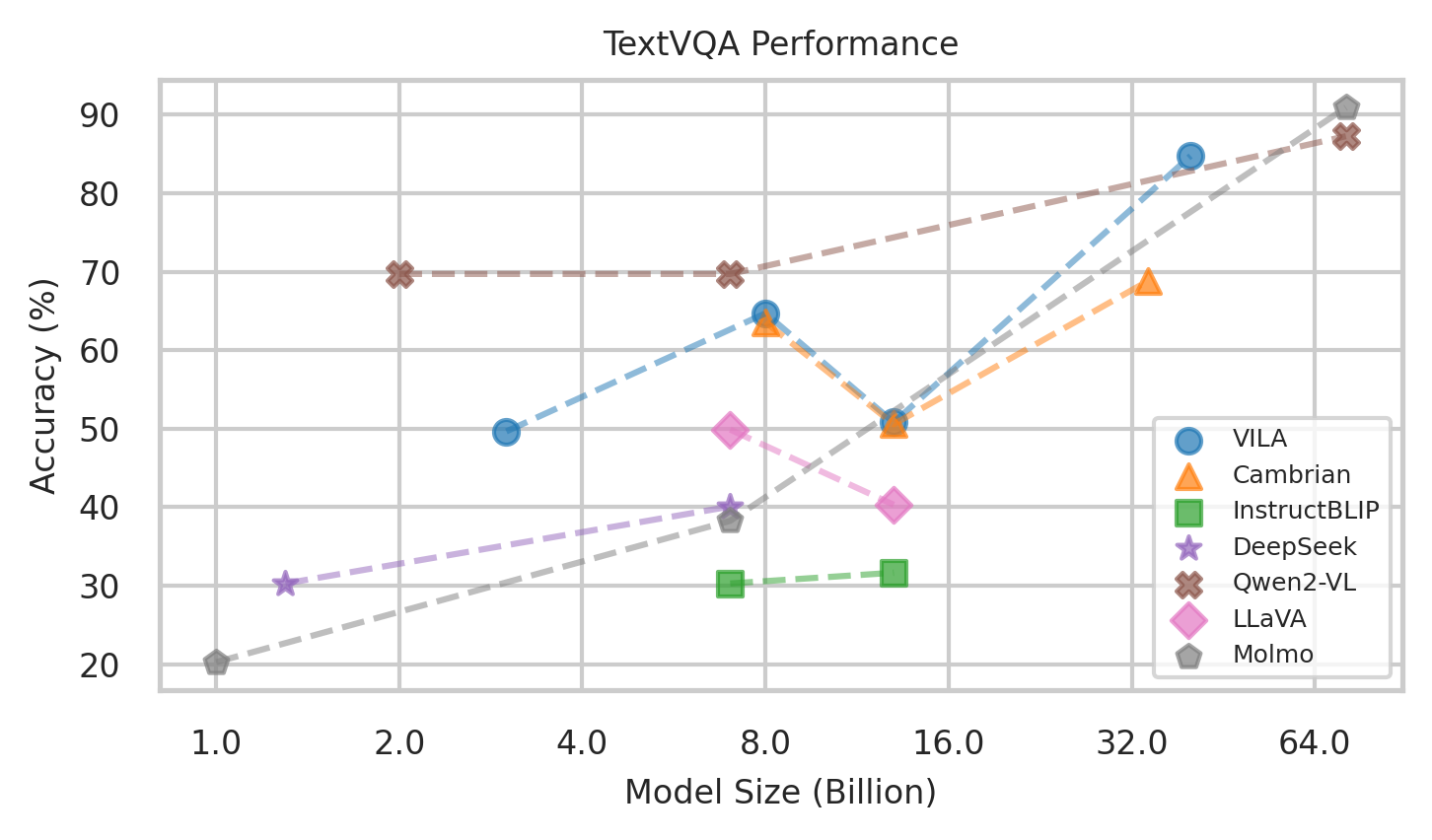}
\includegraphics[width=0.245\textwidth]{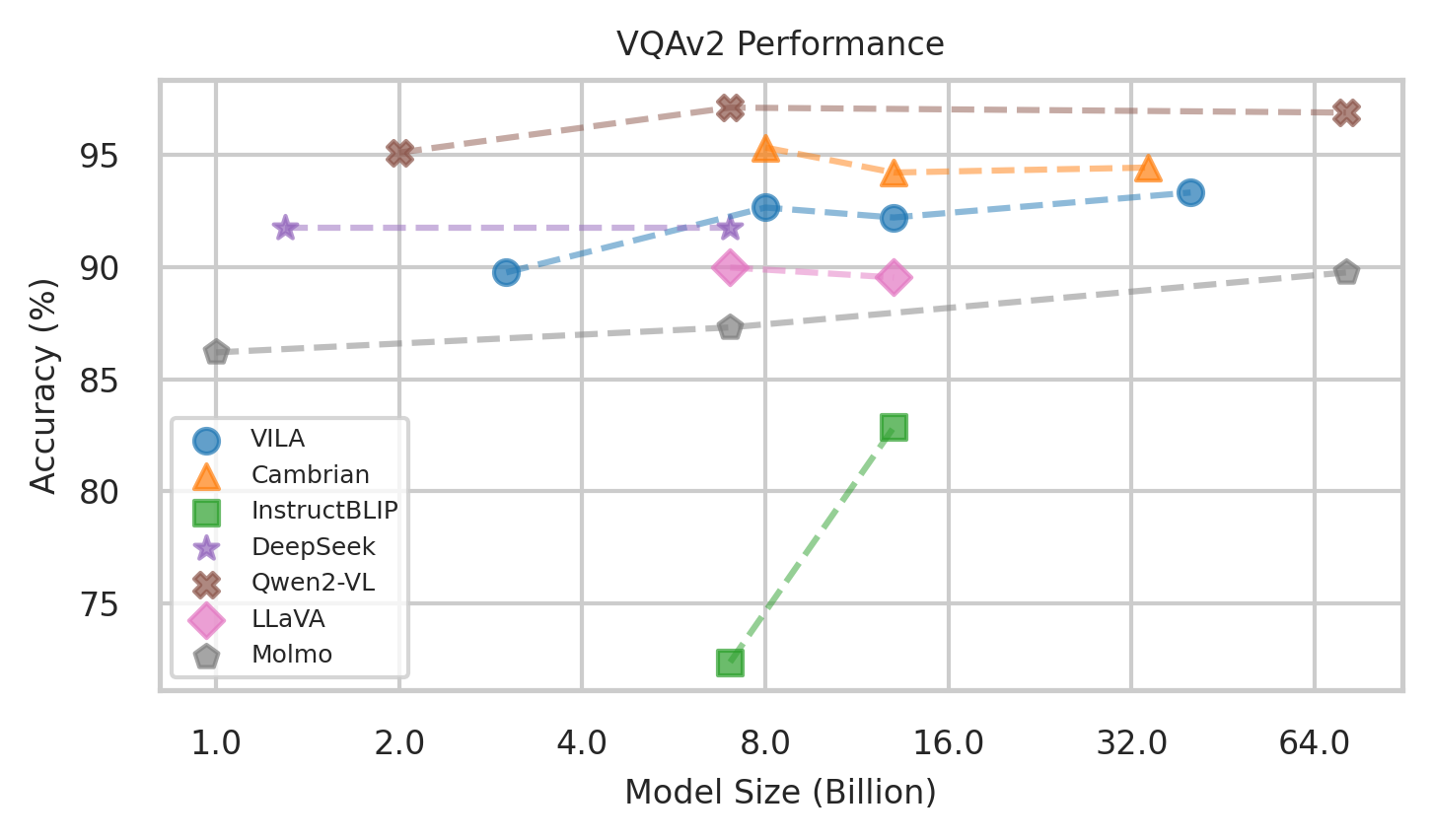}
\includegraphics[width=0.245\textwidth]{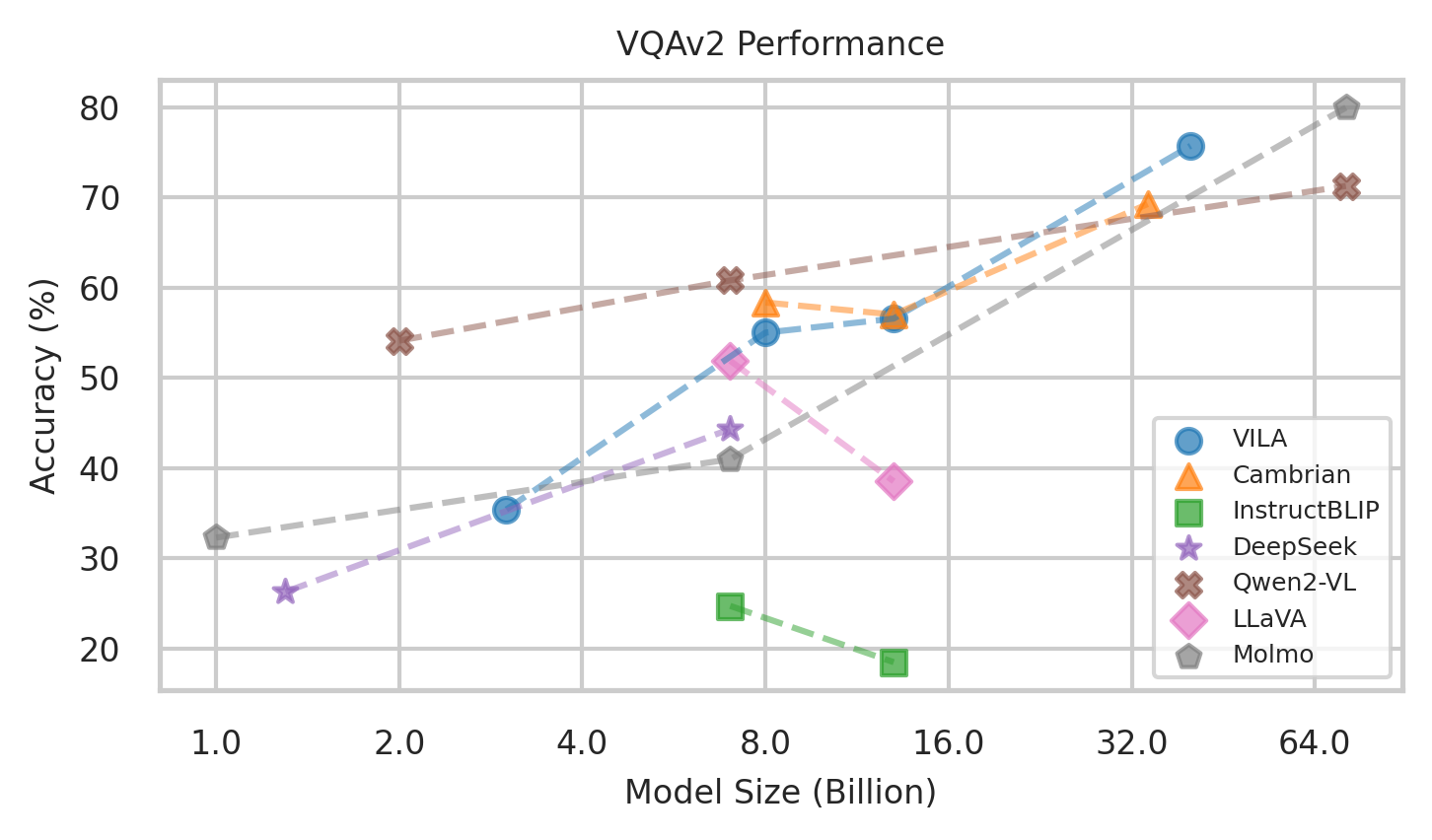}
\includegraphics[width=0.245\textwidth]{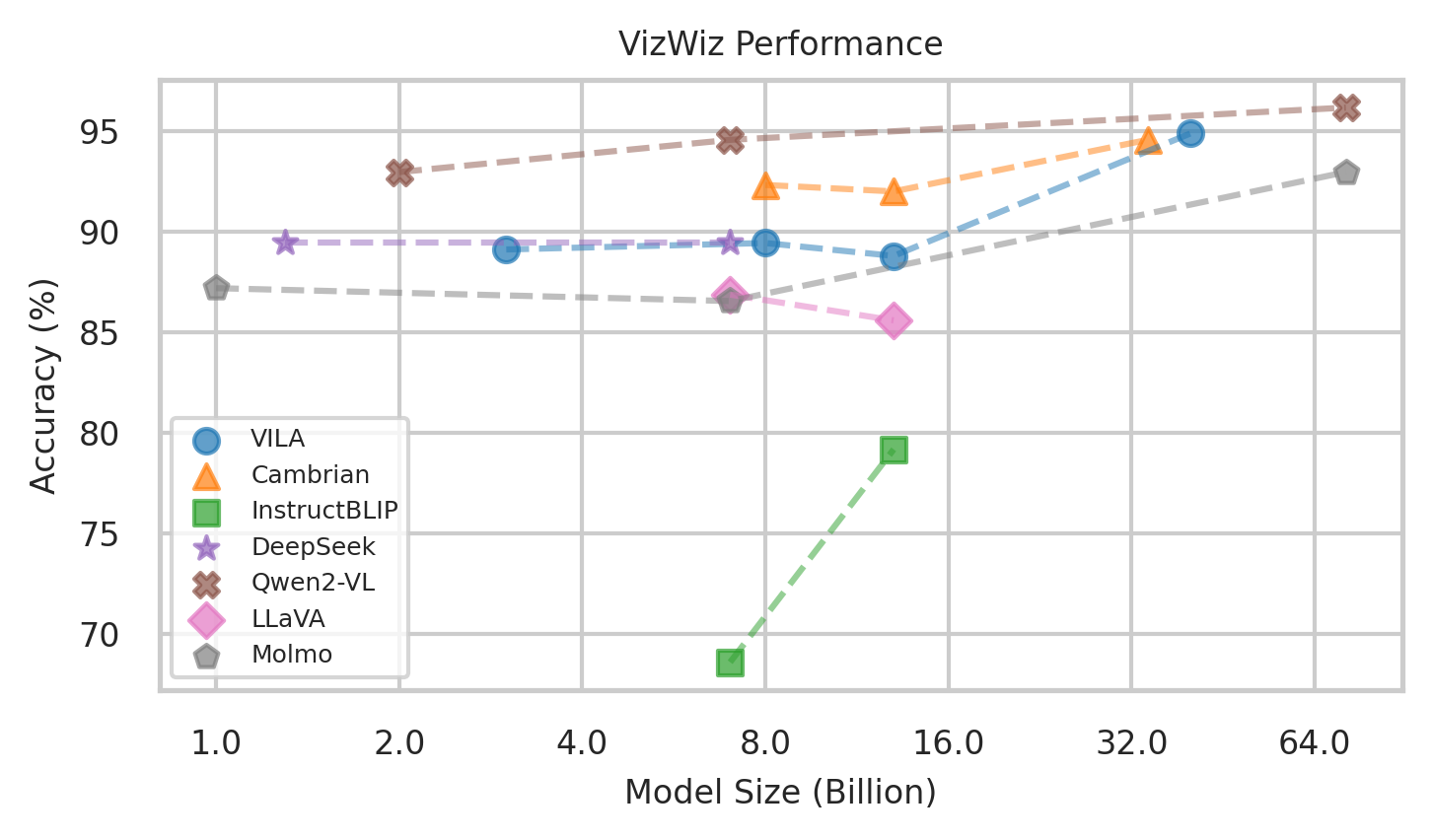}
\includegraphics[width=0.245\textwidth]{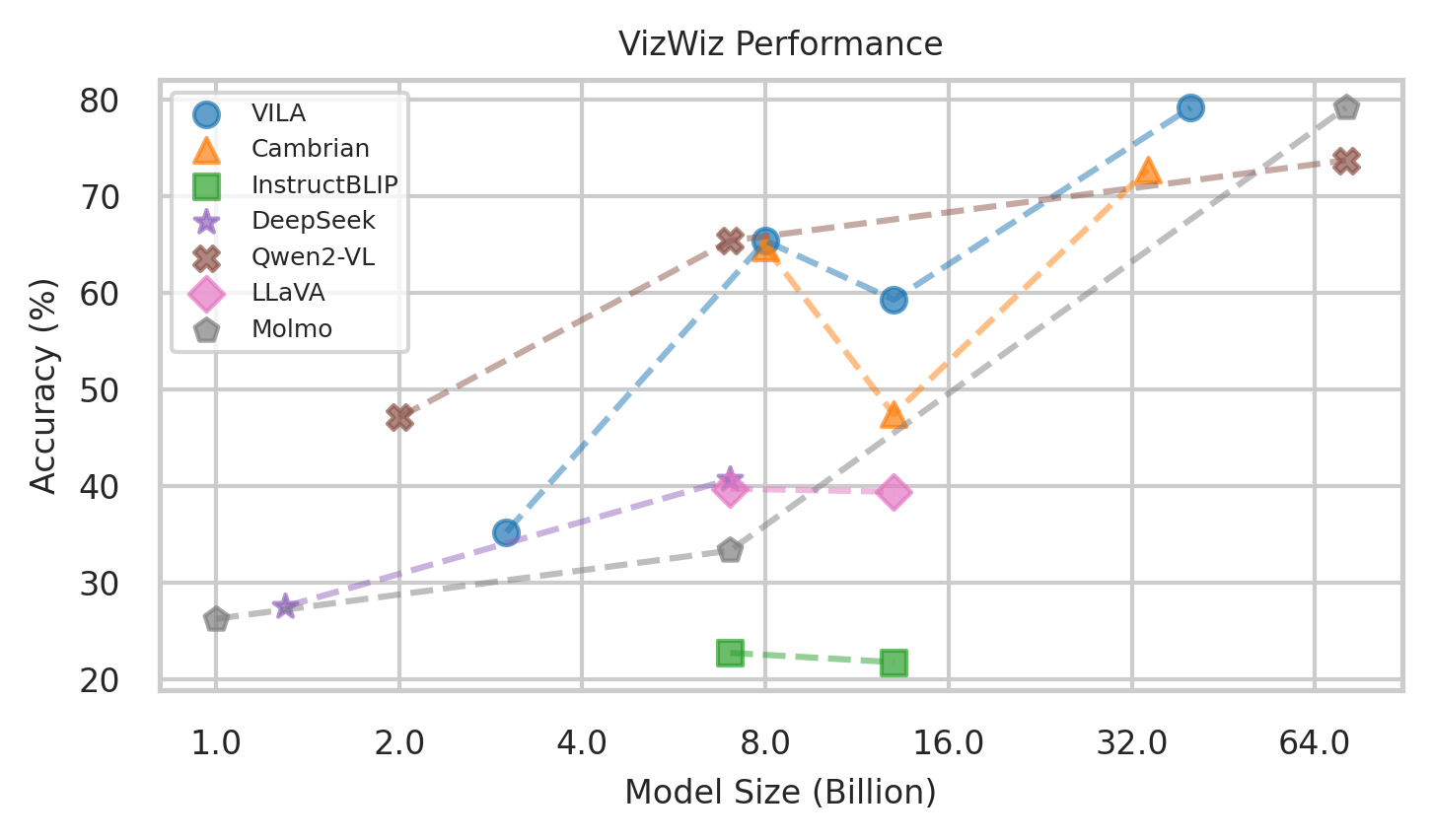}
    \vspace{-2em}
\caption{\textbf{Model performance and scaling analysis on \emph{NegVQA} across different VLM families and datasets.} For each of the 20 subsets in \emph{NegVQA}, we present scaling curves for both the original non-negated dataset and the negated dataset from left to right, resulting in a total of 40 figures.}
    \vspace{-1em}
    \label{fig:scaling_all}
\end{figure*}

\end{document}